\documentclass[10pt,twocolumn,letterpaper]{article}

\usepackage{cvpr}              % To produce the CAMERA-READY version
\definecolor{cvprblue}{rgb}{0.21,0.49,0.74}
\usepackage[pagebackref,breaklinks,colorlinks,allcolors=cvprblue]{hyperref}
\newcommand{\cvtext}[1]{\textcolor{cvprblue}{#1}}

\usepackage{xcolor,colortbl}
\usepackage{amsmath}

\definecolor{Gray}{gray}{0.85}
\definecolor{LightCyan}{rgb}{0.88,1,1}
\newcolumntype{a}{>{\columncolor{Gray}}c}

\usepackage{multirow}
\usepackage{adjustbox}
\usepackage{array}
\usepackage{wrapfig,lipsum}

\newcommand{\yh}[1]{#1}
\newcommand{\jy}[1]{#1}
\newcommand{\hh}[1]{#1}

\usepackage{makecell}

\usepackage{adjustbox}
\usepackage{graphicx}
\usepackage{fix-cm}
\makeatletter
\def\thickhline{%
  \noalign{\ifnum0=`}\fi\hrule \@height \thickarrayrulewidth \futurelet
   \reserved@a\@xthickhline}
\def\@xthickhline{\ifx\reserved@a\thickhline
               \vskip\doublerulesep
               \vskip-\thickarrayrulewidth
             \fi
      \ifnum0=`{\fi}}
\makeatother
\newlength{\thickarrayrulewidth}
\usepackage{setspace}
\usepackage{bm}
\usepackage{tikz}
\newcommand{\tikzxmark}{%
\tikz[scale=0.23] {
    \draw[line width=0.7,line cap=round] (0,0) to [bend left=6] (1,1);
    \draw[line width=0.7,line cap=round] (0.2,0.95) to [bend right=3] (0.8,0.05);
}}
\newcommand{\tikzcmark}{%
\tikz[scale=0.23] {
    \draw[line width=0.7,line cap=round] (0.25,0) to [bend left=10] (1,1);
    \draw[line width=0.8,line cap=round] (0,0.35) to [bend right=1] (0.23,0);
}}
\usepackage{enumitem}

\definecolor{darkgreen}{rgb}{0.2, 0.7, 0.1}

\definecolor{Gray}{gray}{0.8}
\definecolor{LG}{gray}{.92}
\definecolor{bgrey}{RGB}{232,236,250}

\def\paperID{8624}
\def\confName{CVPR}
\def\confYear{2026}

\title{DSERT-RoLL: Robust Multi-Modal Perception for Diverse Driving Conditions with Stereo Event-RGB-Thermal Cameras, 4D Radar, and Dual-LiDAR}

\vspace{-10pt}
\author{
Hoonhee Cho$^{*}$ \qquad
Jae-Young Kang$^{*}$ \qquad
Yuhwan Jeong$^{*}$ \qquad
Yunseo Yang \\
Wonyoung Lee \qquad
Youngho Kim \qquad
Kuk-Jin Yoon \vspace{3pt}\\
{\tt\small Visual Intelligence Lab, KAIST}\\
{\small* equal contribution} \\
{\small\url{https://jeongyh98.github.io/dsert-roll}}
}

\begin{document}
\twocolumn[{%
\renewcommand\twocolumn[1][]{#1}%
\maketitle
\begin{center}
    \vspace{-13pt}
    \centering
    \captionsetup{type=figure}
    \includegraphics[width=.94\textwidth]{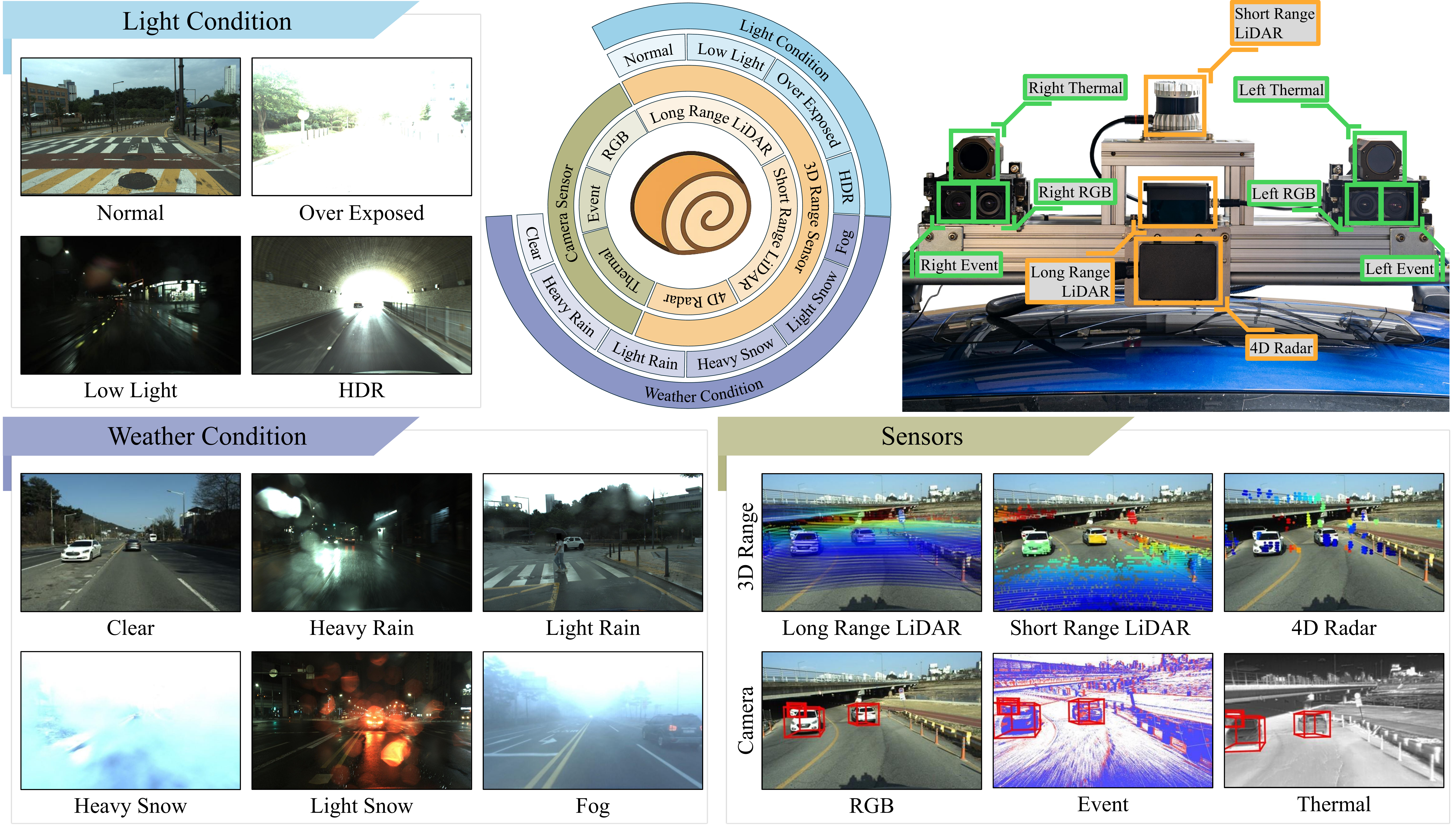}
    \vspace{-6pt}
    \captionof{figure}{The proposed DSERT-RoLL dataset comprises stereo event, RGB, and thermal cameras, together with 4D radar and dual LiDAR, collected in on-road driving across a wide range of weather and illumination conditions, and provided with precise 3D annotations.}
    \label{fig:teaser}
\end{center}%
}
]

\begin{abstract}
In this paper, we present DSERT-RoLL, a driving dataset that incorporates stereo event, RGB, and thermal cameras together with 4D radar and dual LiDAR, collected across diverse weather and illumination conditions. The dataset provides precise 2D and 3D bounding boxes with track IDs and ego vehicle odometry, enabling fair comparisons within and across sensor combinations. It is designed to alleviate data scarcity for novel sensors such as event cameras and 4D radar and to support systematic studies of their behavior. We establish unified 3D and 2D benchmarks that enable direct comparison of characteristics and strengths across sensor families and within each family. We report baselines for representative single modality and multimodal methods and provide protocols that encourage research on different fusion strategies and sensor combinations. In addition, we propose a fusion framework that integrates sensor specific cues into a unified feature space and improves 3D detection robustness under varied weather and lighting. 
\end{abstract}

\section{Introduction}
\label{sec:intro}

\begin{figure*}[t]
    \vspace{-10pt}
    \centering
    \includegraphics[width=.97\textwidth]{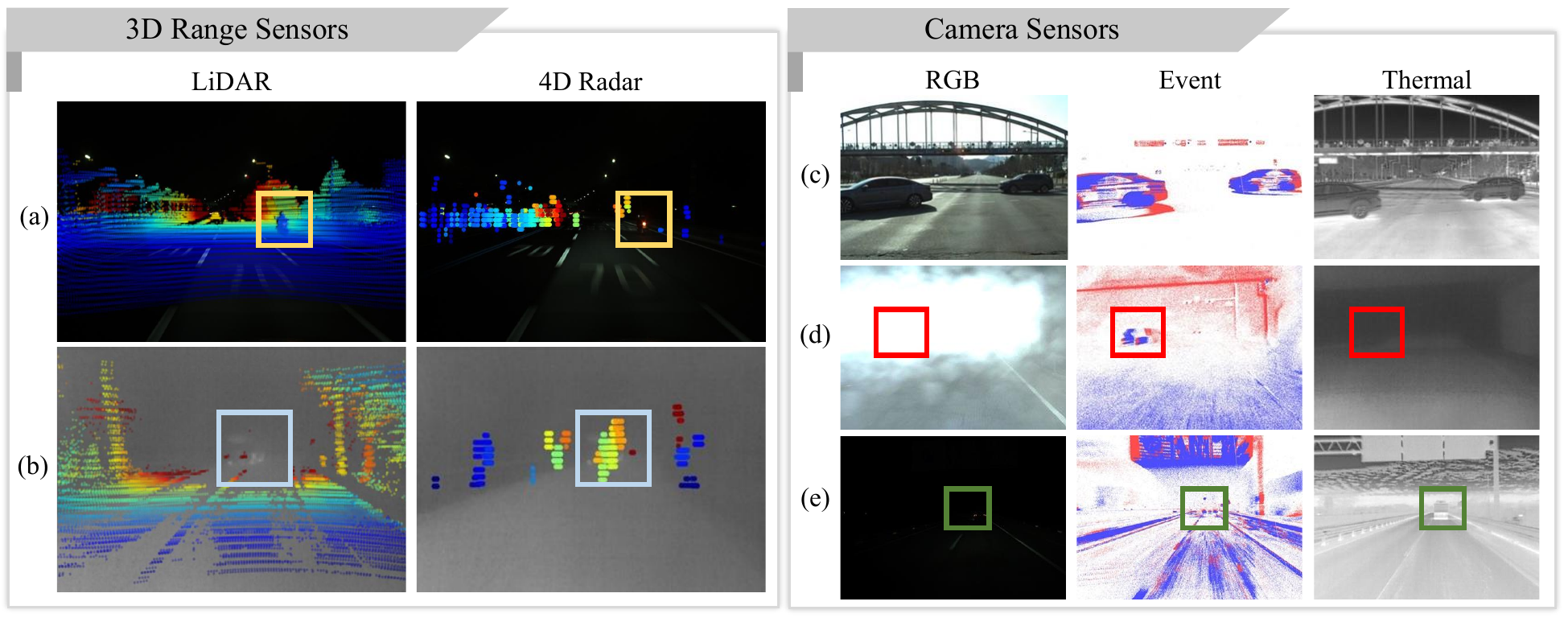}
    \vspace{-9pt}
    \caption{Complementary scenarios across sensor families. (a–b) 3D range sensors: (a) LiDAR-dominant, effective in clear conditions and at long range with accurate geometry; (b) 4D radar-dominant, reliable in adverse weather (\eg,~fog and snow) using Doppler. (c–e) Camera-based sensors: (c) RGB-dominant, strong in daylight and textured scenes; (d) Event-dominant, responsive to small and rapid motions and robust in high dynamic range; (e) Thermal-dominant, informative at night or in low light. Together, (a)–(e) illustrate complementary strengths across sensor types.}
    \label{fig:data_strength}
    \vspace{-10pt}
\end{figure*}

Perception underpins autonomous driving, enabling safe decision-making and control. Early unimodal systems~\cite{zhou2018voxelnet, lang2019pointpillars} suffered from limited accuracy and incomplete scene coverage, motivating multimodal approaches. Fusing cameras and LiDAR~\cite{Yin2024ISFusionIC, liu2022bevfusion} has become standard, combining image semantics with point-cloud geometry to boost robustness and accuracy. Nonetheless, extending these gains to varied weather and lighting remains challenging.

Conventional RGB cameras are widely used in autonomous driving due to their rich semantic information~\cite{tang2024simpb, peng2024learning, yan2024monocd, yang2025bevheight++, liu2024ray, jiang2024far3d}. However, they are sensitive to illumination changes and often degrade under high dynamic range or low light. To mitigate these issues, two alternative sensing modalities have gained attention:

\noindent
\textbf{\textbullet\ Thermal Cameras} operate in the infrared spectrum and are effective in night-time environments where RGB cameras often struggle. They provide complementary cues beyond visible light, thereby enhancing perception robustness under challenging conditions~\cite{shin2023deep, kim2021ms, chen2024tci}.

\noindent
\textbf{\textbullet\ Event Cameras}~\cite{gallego2020event} asynchronously capture changes in brightness at the pixel level, enabling high temporal resolution and low-latency perception. Unlike frame-based cameras, they naturally handle high dynamic range scenes and fast motions, providing complementary information that enhances robustness in challenging environments.

From the perspective of 3D sensing, LiDAR operates independently of illumination but remains sensitive to weather conditions~\cite{behley2019semantickitti, liao2022kitti, hahner2021fog}, where sensing range may decrease, and measurements can become noisy. To mitigate these issues, alternative sensors have been explored.

\noindent
\textbf{\textbullet\ 4D Radar} is a 3D range sensor that emits radio waves and can acquire data even through adverse conditions such as heavy rain or snow. In contrast to LiDAR, which often suffers from reduced range and noise in adverse weather, 4D radar offers more stable perception~\cite{qian2021robust, paek2022k}, serving as a useful complement for all-weather operation.

Research leveraging novel sensors, including event cameras and 4D radar, and, to a lesser extent, thermal cameras, for perception under adverse conditions has accelerated in recent years. However, existing benchmarks~\cite{gehrig2021dsec, paek2022k, choi2018kaist} that include these sensors are typically modality-specific and primarily compare against conventional RGB cameras and LiDAR, leaving direct cross-sensor comparisons and systematic studies of their fusion relatively underexplored.

To advance research in this direction, we present the \textbf{DSERT-RoLL} dataset: \underline{\textbf{D}}riving with \underline{\textbf{S}}tereo \underline{\textbf{E}}vent-\underline{\textbf{R}}GB-\underline{\textbf{T}}hermal Cameras, 4D \underline{\textbf{R}}adar, and Dua\underline{\textbf{L}} \underline{\textbf{L}}iDAR/
As shown in Fig.~\ref{fig:teaser}, we collected multi-modal data across diverse driving scenarios, including night, high dynamic range (HDR), rain, snow, fog, and other challenging conditions. Through DSERT-RoLL, we provide a comprehensive multi-modal benchmark that supports robust perception and fusion studies under diverse driving conditions. Our dataset and approach differ from existing works in the following key aspects:

\begin{table*}[!t]
    \centering
    \caption{Comparison of object detection datasets in autonomous driving. Upper rows use conventional sensors; lower rows include novel sensors. The symbol $\triangle$ marks annotations not officially provided but added by other authors. If 3D boxes exist, ‘Num Data’ counts samples with 3D boxes; otherwise, it counts all samples. Additional comparisons with more datasets are provided in the \textit{supple}.
}
    \vspace{-6pt}
    \setlength\tabcolsep{5.5pt}
    \renewcommand{\arraystretch}{1.0}
\resizebox{1.0\linewidth}{!}{
\begin{tabular}{c||c|cccc|cc|ccc|ccc}
% \hline 
\thickhline 
\multirow{2}{*}{Dataset} &  
Num 
& \multicolumn{4}{c|}{Adverse Weather} & \multicolumn{2}{c|}{3D Range Sensor} 
& \multicolumn{3}{c|}{Camera Sensor}   
& \multicolumn{3}{c}{Ground-truth}   
\\
\cline{3-14}
& Data & Clear & Rain & Fog & Snow & LiDAR & Radar & RGB & Event & Thermal & 3D Bbox. & Tr. ID & Odom\\
\thickhline 
KITTI~\cite{Geiger2012AreWR} & 15k & $\tikzcmark$ & $\tikzxmark$ & $\tikzxmark$ & $\tikzxmark$ & $\tikzcmark$ & $\tikzxmark$ & Stereo & $\tikzxmark$ & $\tikzxmark$ & $\tikzcmark$ & $\tikzcmark$ & $\tikzcmark$ \\
Waymo~\cite{sun2020scalability} & 230k & $\tikzcmark$ & $\tikzcmark$ & $\tikzxmark$ & $\tikzxmark$ & $\tikzcmark$ & $\tikzxmark$ & Multi-view & $\tikzxmark$ & $\tikzxmark$ & $\tikzcmark$ & $\tikzcmark$ & $\tikzxmark$\\
NuScenes~\cite{caesar2020nuscenes} & 40k & $\tikzcmark$ & $\tikzcmark$ & $\tikzxmark$ & $\tikzxmark$ & $\tikzcmark$ & 3D & Multi-view & $\tikzxmark$ & $\tikzcmark$ & $\tikzcmark$ & $\tikzcmark$ & $\tikzcmark$\\
Argoverse 2~\cite{wilson2023argoverse} & 150k & $\tikzcmark$ & $\tikzcmark$ & $\tikzxmark$ & $\tikzcmark$  & $\tikzcmark$ & $\tikzxmark$ & Multi-view & $\tikzxmark$ & $\tikzxmark$ & $\tikzcmark$ & $\tikzcmark$ & $\tikzcmark$ \\
\hline
K-Radar~\cite{paek2022k} & 35k & $\tikzcmark$ & $\tikzcmark$ & $\tikzcmark$ & $\tikzcmark$ & $\tikzcmark$ & 4D & Multi-view & $\tikzxmark$ & $\tikzxmark$ & $\tikzcmark$ & $\tikzcmark$ & $\tikzcmark$ \\
TJ4DRadSet~\cite{zheng2022tj4dradset} & 7.8k & $\tikzcmark$ & $\tikzxmark$ & $\tikzxmark$ & $\tikzxmark$ & $\tikzcmark$ & $\tikzcmark$ & Mono & $\tikzxmark$ & $\tikzxmark$ & $\tikzcmark$  & $\tikzcmark$ & $\tikzcmark$ \\
DSEC~\cite{gehrig2021dsec} & 5.4k & $\tikzcmark$ & $\tikzxmark$ & $\tikzxmark$ & $\tikzxmark$ & $\tikzcmark$ & $\tikzxmark$  & Stereo & Stereo & $\tikzxmark$ & $\triangle$ & $\triangle$ & $\tikzcmark$ \\
1Mpx~\cite{perot2020learning} & 32M & $\tikzcmark$ & $\tikzcmark$ & $\tikzxmark$ & $\tikzxmark$ & $\tikzxmark$ & $\tikzxmark$  & Mono & Mono & $\tikzxmark$ & $\tikzxmark$ & $\tikzxmark$ & $\tikzxmark$  \\
SeeingThroughFog~\cite{bijelic2020seeing} & 13.5k & $\tikzcmark$ & $\tikzcmark$ & $\tikzcmark$ & $\tikzcmark$ & $\tikzcmark$ & 3D & Stereo & $\tikzxmark$ & Mono & $\tikzcmark$ & $\tikzxmark$  & $\tikzxmark$ \\
KAIST~\cite{choi2018kaist} & 8.9k & $\tikzcmark$ & $\tikzxmark$ & $\tikzxmark$ & $\tikzxmark$ & $\tikzcmark$ & $\tikzxmark$ & Stereo & $\tikzxmark$ & Mono & $\tikzxmark$ & $\tikzxmark$ & $\tikzxmark$ \\
\rowcolor{bgrey}
\textbf{DSERT-RoLL (Ours)} & 22k & $\tikzcmark$ & $\tikzcmark$ & $\tikzcmark$ & $\tikzcmark$ & $\tikzcmark$ & 4D & Stereo & Stereo & Stereo & $\tikzcmark$ & $\tikzcmark$  & $\tikzcmark$ \\
\thickhline
% \bottomrule
\end{tabular}
}
\label{tab:dataset_compare}
\vspace{-8pt}
\end{table*}

\begin{itemize}[noitemsep, topsep=0pt]
\item While recent datasets have analyzed the advantages of novel sensors, the evaluation of various sensors in the same environment remains largely unexplored. In contrast, the DSERT-RoLL dataset includes widely researched emerging sensors and data collected from extreme environments. By providing a fair benchmark for training and evaluation across multiple sensors in the same setting, DSERT-RoLL enables a deeper analysis of each sensor's strengths and characteristics.

\item Although there are numerous benchmarks and datasets based on widely used sensors like frame cameras and LiDAR, benchmarks based on emerging sensors, such as event cameras, thermal cameras, and 4D radar, are relatively scarce. The proposed DSERT-RoLL dataset contributes to enhancing data richness and is expected to support a wide range of studies, including domain adaptation, domain generalization, and more.

\item 
While differences between camera sensors and 3D range sensors are well documented, Fig.~\ref{fig:data_strength} shows that complementary strengths also exist within each sensor type, for example among cameras such as RGB, event, and thermal, and among 3D range sensors such as LiDAR and 4D radar. Each type excels under different failure modes, which suggests synergy rather than simple substitution. Building on this, we propose a framework that leverages these complementary strengths to achieve robust 3D object detection under varying weather and illumination. By integrating cues into a unified feature space, our method improves perception reliability across conditions.
\end{itemize}

\section{Related Works}
\label{sec:related_works}

\noindent
\textbf{Traditional Driving Datasets.}
Numerous datasets~\cite{Geiger2012AreWR, wilson2023argoverse, caesar2020nuscenes, sun2020scalability, cordts2016cityscapes} have demonstrated the potential for safe autonomous driving perception~\cite{li2024fully,xia2024hinted, chang2024unified, zhang2024general} through the use of RGB and LiDAR-based data~\cite{pham20203d, xiao2021pandaset, mao2021one}, human annotations, and extensive training with large-scale datasets. Subsequent research~\cite{pitropov2021canadian, diaz2022ithaca365} has incorporated more diverse driving scenes under various conditions, and by expanding the scale of data, it has enabled the development of more robust perception algorithms.
However, due to the inherent limitations of RGB and LiDAR sensors, their robustness in extreme environments (\eg,~fog, snow) remains insufficient. Consequently, research in this area is advancing with the emergence of novel sensors~\cite{bijelic2020seeing, palladin2024samfusion, zhang2023delivering, zheng2024learning}, fostering additional fusion studies.

\noindent
\textbf{Thermal Camera-based Driving Datasets.} 
Thermal imaging captures emitted infrared radiation rather than reflected visible light, providing illumination-invariant cues that complement RGB, especially in darkness and adverse weather. This is particularly valuable for object detection in road scenes, where nighttime and inclement conditions demand robust perception across diverse environments.
Recent RGB–thermal benchmarks~\cite{shin2023deep, shin2025deep, choi2018kaist, FLIR_ADAS_Dataset, hwang2015multispectral, gonzalez2016pedestrian} have catalyzed research on multimodal detection, segmentation~\cite{kutuk2022semantic, ha2017mfnet}, and tracking~\cite{berg2015thermal, el2025thermal} under challenging illumination and weather~\cite{bijelic2020seeing, takumi2017multispectral,liu2022target}. Alongside dataset growth, fusion methodologies have matured from early feature concatenation to more principled designs~\cite{devaguptapu2019borrow, el2023enhanced, munir2021sstn, xiao2024gm, zhou2020improving, li2018multispectral, jang2025multispectral, vs2022meta}.

\noindent
\textbf{Event Camera-based Driving Datasets.}
Event cameras asynchronously report per-pixel brightness changes with microsecond latency and extremely high dynamic range, producing motion-blur-free signals~\cite{cho2024benchmark, kim2025sharp} that complement frame-based RGB and LiDAR under fast ego-motion~\cite{hidalgo2022event, mueggler2017event, burner2022evimo2, mitrokhinev}, low light~\cite{kim2024towards, liu2024seeing, xia2023cmda, li2024event, yao2025event, cho2024finding}, and glare~\cite{zou2024eventhdr, mostafavi2021learning}. These properties make them well suited for autonomous-driving perception in road scenes, where rapid maneuvers and abrupt illumination transitions demand temporally precise, HDR sensing. Recent event-driven driving benchmarks with single~\cite{de2020large, binas2017ddd17} or stereo sensors~\cite{gehrig2021dsec, peng2024cosec, zhu2018multivehicle, cho2025ev}, aligned RGB cameras, and vehicle telemetry have supported progress across a broad range of perception tasks~\cite{cho2025ev, perot2020learning, kang2025temporal, Chaney_2023_CVPR, cho2023learning, cho2023non, gehrig2021raft, rebecq2019high, wang2024event, sun2022ess, jeong2024towards, kang2025unleashing}. 
Despite these advantages, multimodal fusion beyond fusion with RGB~\cite{zhao2024edge, cho2022selection, cho2022event} remains underexplored in event-based vision, and annotations for 3D perception are still scarce.
By providing additional modalities alongside events and 3D annotations, this work serves as a strong foundation for subsequent research on event cameras.

\begin{figure*}[t]
    \centering
    \includegraphics[width=.978\textwidth]{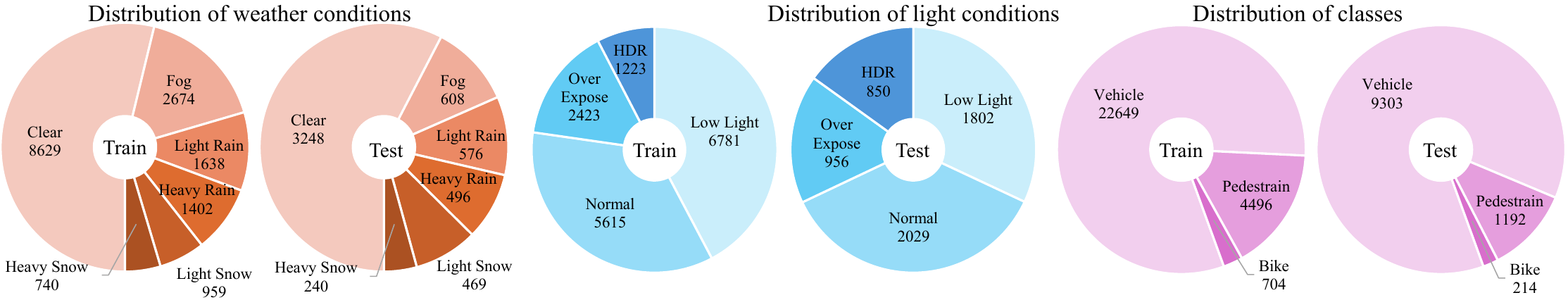}
    \vspace{-8pt}
    \caption{Distribution of training and testing data with respect to weather conditions, lighting conditions, and object classes.}
    \label{fig:data_statistic}
    \vspace{-10pt}
\end{figure*}

\noindent
\textbf{4D Radar-based Driving Datasets.}
Automotive 4D radar measures range, azimuth, elevation, and radial velocity (Doppler), delivering long-range, illumination- and weather-robust cues that complement RGB and LiDAR with direct motion observables~\cite{fan20244d}. 
These properties are attractive for perception~\cite{paek2022k, yang2024v2x} in adverse conditions, such as fog, rain, snow, and nighttime, where temporally stable velocity estimates and extended detection ranges are critical. Recent 4D radar driving datasets~\cite{zhang2021raddet, alibeigi2023zenseact, zheng2022tj4dradset, zhang2021raddet, schumann2021radarscenes, palffy2022multi, huang2025v2x} provide synchronized radar point clouds together with LiDAR and cameras, and have demonstrated successful perception under these adverse conditions.
As LiDAR datasets have grown, a wide range of sensor combination studies have emerged. By contrast, 4D radar is relatively recent and has primarily been paired with cameras or LiDAR. Our work aims to provide a foundation for exploring richer 4D radar-based multimodal configurations, including event cameras and thermal cameras, enabling broader research on these combinations.

\noindent
\textbf{Multi-modal 3D Object Detection.}
3D object detection~\cite{lin2024monotta,ji2024enhancing,park2024selectively,wang2024towards} aims to estimate 3D bounding boxes and object orientations. Unimodal LiDAR approaches~\cite{zhang2024voxel, chen2023focalformer3d, zhang2024safdnet, liu2024lion, jin2025geoformer, liu2024seed,li2024bevnext} leverage the depth accuracy of point clouds to regress 3D boxes. Recently, multimodal fusion has been actively explored to exploit the complementary strengths of different sensors under diverse conditions. The most common setup fuses RGB images with LiDAR~\cite{yin2024fusion, wang2025mv2dfusion, fan2025mgaf, chen2023futr3d}. This pairing adds color and texture to precise depth and improves small object recall and 3D localization. To improve robustness in adverse weather and low light, radar~\cite{lin2024rcbevdet, zhou2023bridging,hwang2022cramnet} has also been incorporated. With the advent of a 4D radar that provides range, Doppler, azimuth, and elevation information, camera and radar fusion~\cite{chae2025doppler, chae2024towards} has advanced further. Moreover, recent architectures ~\cite{palladin2024samfusion, chen2023futr3d} enable fusion of two or more modalities within a unified framework.

\section{DSERT-RoLL Dataset}
\label{sec:dataset}

\subsection{Sensor Configuration}

\begin{table}[t]
\centering
\setlength\tabcolsep{5.1pt}
\caption{Sensor suit details.}
\vspace{-7pt}
\resizebox{.478\textwidth}{!}{
\begin{tabular}{lccccc}
% \hline
\thickhline
% Dataset & Modality  & Events 
\rowcolor{LG}
Sensors& Model Name & Resolution & FoV & FPS
\\ 
% \hline
\thickhline
RGB & 2 $\times$ BFS-U3-51S5C
& 2448 $\times$ 2048 & 82.2$^{\circ}$ $\times$ 66.5$^{\circ}$ & 10 \\
\hline
Event & 2 $\times$ Prophesee EVK4 & 1280 $\times$ 720 & 76.7$^{\circ}$ $\times$ 65.5 $^{\circ}$	 & $>$10k \\
\hline
Thermal & 2 $\times$ FLIR A65 & 640 $\times$ 512 & 90$^{\circ}$ $\times$ 69$^{\circ}$ & 30 \\
\thickhline
\rowcolor{LG}
%  &  &   Max  & &  \\
% \rowcolor{LG}
% \multirow{-2}{*}{Sensors}&\multirow{-2}{*}{Model Name} & Range &  \multirow{-2}{*}{FoV} & \multirow{-2}{*}{FPS}
% \\
Sensors & Model Name & Max. Range & FoV & FPS \\
\thickhline
4D Radar & RETINA-4FN &  100m & 100$^{\circ}$ $\times$ 24$^{\circ}$ & 20\\
\hline
Long-range & \multirow{2}{*}{Livox HAP}  
 &  \multirow{2}{*}{150m} &  \multirow{2}{*}{120$^{\circ}$ $\times$ 25$^{\circ}$} & \multirow{2}{*}{10}\\
LiDAR \\
\hline
Short-range & \multirow{2}{*}{os0-128} & \multirow{2}{*}{100m} & \multirow{2}{*}{360$^{\circ}$ $\times$ 90$^{\circ}$} & \multirow{2}{*}{20}\\
LiDAR \\
\hline
\multirow{2}{*}{GPS/IMU} & Microstrain & \multirow{2}{*}{N/A} & \multirow{2}{*}{N/A} & \multirow{2}{*}{10/100} \\
% IMU \\ 
& 3DM-GX5-45 \\
\thickhline
% \hline
\label{tab:dataset_comparison}
\vspace{-25pt}
\end{tabular}}
\end{table}

As illustrated in Fig.~\ref{fig:teaser}, we equipped the vehicle with the multi-modal sensor setting described in Table~\ref{tab:dataset_comparison}.
We first mounted LiDAR sensors, which are widely adopted 3D range sensors for object detection, including a long-range LiDAR and a high-resolution short-range LiDAR. The long-range LiDAR is used to obtain reliable object annotations over extended distances, whereas the short-range LiDAR provides high-resolution fine-grained point measurements with an enlarged vertical field of view, thereby offering dense geometric coverage. To further ensure perception performance under extreme weather such as fog or snow, we additionally equipped the vehicle with a 4D Radar. All cameras were deployed in stereo configurations to fully cover the frontal field of view, and we extended the setup beyond RGB sensors by incorporating a thermal camera and an event camera. These complementary modalities enhance robustness against challenging lighting conditions and motion blur, providing reliable perception in adverse environments. Finally, a GPS antenna and IMU sensors were mounted near the camera suite on the vehicle to enable precise localization of the ego-vehicle.

\begin{figure*}[t]
    \centering
    \includegraphics[width=1.0\textwidth]{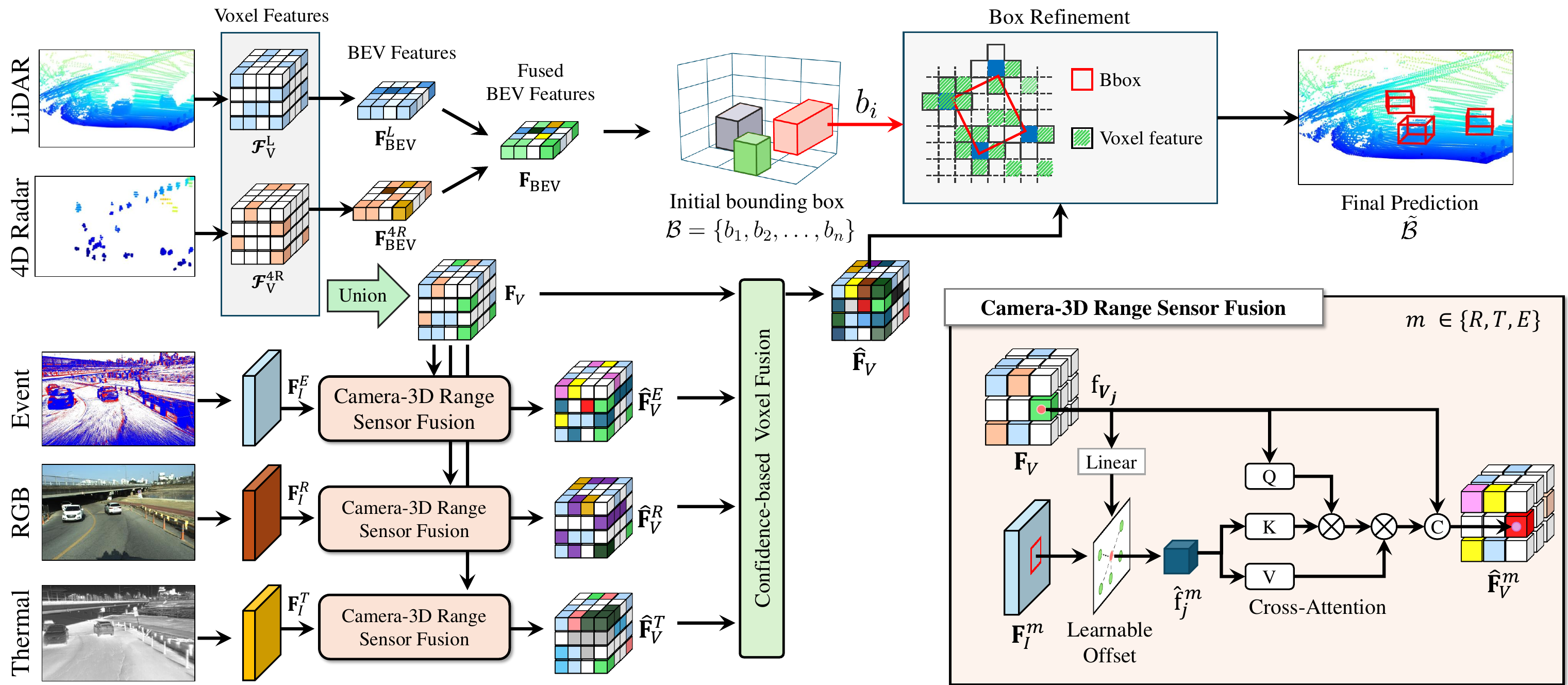}
    \vspace{-14pt}
    % \caption{Overall framework of the proposed multi-modal method.}
    \caption{Overview of the proposed multi-modal 3D detection framework. LiDAR and 4D Radar features are voxelized and fused to generate initial 3D box proposals. RGB, thermal, and event features are then projected into 3D space via voxel-centric sampling and integrated through confidence-based fusion. The refined fused features are used for final bounding box prediction.}
    \label{fig:overall_method}
    \vspace{-10pt}
\end{figure*}

\subsection{Data Distribution}
The DSERT-RoLL dataset encompasses diverse driving scenarios, including highways, urban streets, suburban roads, and narrow alleys. In total, the data collection process results in 22K frames of multi-modal sensor measurements captured under various environmental conditions. 

The DSERT-RoLL dataset is categorized into six weather conditions: clear, fog, light rain, heavy rain, light snow, and heavy snow. This design allows a fair comparison of the strengths of different sensor modalities under diverse weather conditions. A key advantage of DSERT-RoLL is that multiple modalities are captured simultaneously in the same environment, enabling deeper research on multi-sensor fusion, particularly for 3D range sensors.

While 3D range sensors (\eg,~LiDAR and Radar) are largely unaffected by illumination, camera-based perception can vary significantly depending on lighting. To facilitate fusion research in such scenarios, we further categorize the data into four lighting conditions: normal, low light, overexposed, and HDR.

The dataset defines three object categories, namely \textit{vehicle}, \textit{pedestrian}, and \textit{bike}, which represent the most common types of driving datasets. We split the dataset into training and test sets with a 7:3 ratio. As shown in Fig.~\ref{fig:data_statistic}, the distributions of weather, lighting, and object classes are balanced across both splits. This balanced coverage across diverse scenarios and environmental conditions makes DSERT-RoLL a reliable benchmark for evaluating the robustness of perception models in challenging real-world settings.

\section{Multi-modal Approach on 3D Detection}
\label{sec:method}

To demonstrate the benefits of leveraging multi-modality, we propose a method that effectively fuses diverse sensor modalities into a unified feature space. This in-depth methodology is made possible by the strength of our DSERT-RoLL benchmark dataset, which incorporates multiple modalities within a single collection. In this section, we elaborate on the design and implementation of the proposed approach and highlight how multi-modal fusion contributes to robust perception performance.

\noindent
\textbf{Overall Framework.}
% As shown in Fig.~\ref{fig:overall_method}, 
As illustrated in Fig.~\ref{fig:overall_method}, the inputs to our framework consist of 3D range sensors and multi-modal single-view images. For the 3D range sensor, we use a LiDAR (\(L\)) and a 4D Radar (\(4R\)) providing point sets \(\mathcal{P}^L = \{(x_i, y_i, z_i) | f_i^L\}_{i=1}^{N_L}\) and \(\mathcal{P}^{4R} = \{(x_j, y_j, z_j) |  f_j^{4R}\}_{j=1}^{N_{4R}}\). Here, \((x, y, z)\) denotes the 3D spatial coordinate of each point, and \(f \in \mathbb{R}^{C_p}\) represents point-wise features such as intensity (LiDAR) or Doppler velocity (4D Radar). \(N_L\) and \(N_{4R}\) denote the number of points from LiDAR and 4D Radar, respectively. 
Each point cloud is processed by a 3D voxel-based backbone~\cite{zhou2018voxelnet} to obtain voxel features $\mathcal{F}_V^L$ and $\mathcal{F}_V^{4R}$, with $\mathcal{F}_V \in \mathbb{R}^{X \times Y \times Z \times C_V}$, where $C_V$ denotes the number of channels and $(X, Y, Z)$ represents the voxel grid size. 
For the camera sensors, we incorporate three modalities: an RGB image $\mathbf{I}^R \in \mathbb{R}^{H \times W \times 3}$, a thermal image $\mathbf{I}^T \in \mathbb{R}^{H \times W \times 1}$, and a voxel grid~\cite{zhu2019unsupervised} of events $\mathbf{I}^E \in \mathbb{R}^{H \times W \times 5}$, which are processed by a 2D backbone~\cite{Liu2021SwinTH} to extract features $\mathbf{F}_I^R, \mathbf{F}_I^T, \mathbf{F}_I^E \in \mathbb{R}^{H/4 \times W/4 \times C_I}$, where $C_I$ denotes the number of channels for RGB, thermal, and event features.

We first generate initial 3D bounding boxes from the range sensors and then refine them by incorporating complementary cues from the camera sensors, where confidence is taken into account during the fusion. Finally, all heterogeneous inputs are processed through our proposed multi-modal fusion strategy to produce unified representations for 3D object detection.

\noindent
\textbf{Initial 3D Box Proposal from Range Sensors.}
To perform effective and efficient computation in 3D space, the voxel features $\mathcal{F}_V^L$ and $\mathcal{F}_V^{4R}$ are collapsed along the vertical axis and transformed by 2D convolutions on the ground plane into bird’s-eye-view (BEV) features $\mathbf{F}_{\mathrm{BEV}}^L, \mathbf{F}_{\mathrm{BEV}}^{4R} \in \mathbb{R}^{\tfrac{X}{s} \times \tfrac{Y}{s} \times C_{B}}$, where $X$ and $Y$ are the voxel grid dimensions in the horizontal plane, $s$ is the stride, and $C_{B}$ is the channel dimension. 
Given the BEV features $\mathbf{F}_{\mathrm{BEV}}^L$ and $\mathbf{F}_{\mathrm{BEV}}^{4R}$, 
we concatenate the two modality features along the channel dimension and apply a convolutional layer to fuse them. This simple yet effective fusion produces cross-modally enriched BEV representations while maintaining computational efficiency.
Through the detector~\cite{yin2021center}, we obtain the initial set of bounding boxes $\mathcal{B} = \{ b_1, b_2, \ldots, b_n \}$, where $n$ is a pre-defined box number.

\setlength{\aboverulesep}{-1.5pt}
\setlength{\belowrulesep}{0pt}
\setlength{\tabcolsep}{11.5pt}
\renewcommand{\arraystretch}{1.0}
\begin{table*}[t]
\begin{center}
\caption{Ablation study across sensor modalities on the DSERT-RoLL dataset for 3D detection. For the modalities, we use the following notation: R: RGB, E: Event, T: Thermal, 4R: 4D Radar, and L: LiDAR.
}
\vspace{-7pt}
\label{tab:main_modality}
\resizebox{.99\linewidth}{!}{
\begin{tabular}{c|cccccc|cccc}
\toprule
\multirow{3}{*}{Modality}& \multicolumn{6}{c|}{Weather Condition}  & 
\multicolumn{4}{c}{Light Condition} \\
\cline{2-11}
& \multirow{2}{*}{Clear} & \multirow{2}{*}{Fog} & Light & Heavy & Light & Heavy  & \multirow{2}{*}{Normal} & Low  & Over  & \multirow{2}{*}{HDR} \\
& & & Rain & Rain & Snow & Snow & & Light & Expose & 
\\
\hline
L & 82.90 & 65.67 & 89.62 & 62.97 & 77.26 & 54.14 & 74.71 & 86.10 & 75.82 & 74.51\\
R+L & 84.67 & 66.14 & 90.29 & 72.82 & 78.40 & 59.43 & 76.26 & 87.41 & 77.55 & 79.31 \\
4R+L & 88.26 & 67.41 & 91.43 & 67.41 & 77.43 & 69.96 & 79.43 & 88.73 & 82.85 & 82.98\\
R+4R+L & 88.35 & 67.38 & 91.79 & 79.03 & 84.39 & 70.26 & 81.31 & 91.04 & 80.34 & 83.93 \\
R+E+4R+L & 88.70 & \textbf{71.45} & 92.94 & 80.11 & 82.75 & 71.64 & 81.92 & 91.43 & 83.37 & \textbf{86.55}\\
R+T+4R+L & 89.48 & 71.00 & 93.94 & 79.77 & 84.02 & 71.32 & 82.26 & 92.20 & 83.20 & 85.66 \\
\rowcolor{bgrey} R+E+T+4R+L & \textbf{90.30} & 71.42 & \textbf{95.10} & \textbf{80.26} & \textbf{85.59} & \textbf{72.94} & \textbf{82.93} & \textbf{92.65} & \textbf{85.47} & 86.33\\
\bottomrule
\end{tabular}
}
\end{center}
\vspace{-20pt}
\end{table*}

% \cite{yin2021center}

\noindent
\textbf{Camera-3D Range Sensor Fusion.} 
We aim to leverage the camera features, which contain multiple strengths and rich semantic information, to gain additional performance. To integrate image features into the 3D space in a multi-modal manner, we propose a voxel-centric sampling strategy. Specifically we extract non-empty voxel indices from the LiDAR and 4D Radar voxel features $\mathcal{F}_V^{L}$ and $\mathcal{F}_V^{4R}$ to obtain $\mathbf{F}_V^{L}$ and $\mathbf{F}_V^{4R}$. We then combine the features on the union of these indices to form a unified sparse voxel feature space.
 We denote the resulting set as
\begin{equation}
\mathbf{F}_V \;=\; \mathbf{F}_V^L \cup \mathbf{F}_V^{4R}
\;=\; \{(V_j, \mathbf{f}_{V_j})\}_{j=1}^{N_V},
\end{equation}
where $V_j$ is the voxel index, $\mathbf{f}_{V_j}\in\mathbb{R}^{C_V}$ is the fused feature of a non-empty voxel, and $N_V$ is the number of non-empty voxels. Let $\Omega^L$ and $\Omega^{4R}$ denote the sets of non-empty voxels for the LiDAR and 4D Radar, respectively.
For each $V_j$ in the union $\Omega=\Omega^L\cup\Omega^{4R}$, the fused feature is defined by
\begin{equation}
\mathbf{f}_{V_j} =
\begin{cases}
\mathbf{f}^{L}_{V_j} & \text{if } V_j \in \Omega^L\setminus\Omega^{4R},\\[2pt]
\mathbf{f}^{4R}_{V_j} & \text{if } V_j \in \Omega^{4R}\setminus\Omega^L,\\[2pt]
\mathbf{P}\!\bigl[\mathbf{f}^{L}_{V_j}~|\,\mathbf{f}^{4R}_{V_j}\bigr] & \text{if } V_j \in \Omega^L\cap\Omega^{4R},
\end{cases}
\end{equation}
where $[\cdot~|~\cdot]$ denotes channel-wise concatenation and $\mathbf{P}$ is a per-scale linear projector that maps the concatenated $2C_V$ channels back to $C_V$. Thus, voxels with neither modality remain absent, voxels with exactly one modality keep that feature as-is, and voxels with both modalities are concatenated and projected to preserve dimensionality while enabling cross-modal fusion.

For each non-empty voxel $V_j$, we obtain modality-specific projections onto the image planes of the RGB, thermal, and event cameras as
$
u_j^R = \mathbf{M}^R \cdot V_j, \quad
u_j^T = \mathbf{M}^T \cdot V_j, \quad
u_j^E = \mathbf{M}^E \cdot V_j,
$
where $\mathbf{M}^R$, $\mathbf{M}^T$, and $\mathbf{M}^E$ are the projection matrices, which are the products of the intrinsic and extrinsic matrices for each modality. The projected locations $u_j^R, u_j^T, u_j^E$ are used to sample nearby image features from $\mathbf{F}_I^R$, $\mathbf{F}_I^T$, and $\mathbf{F}_I^E$, respectively. 
Given the modality-specific projections $u_j^R, u_j^T, u_j^E$, aggregated image features are obtained by sampling feature values in the neighborhood of each projection. 
% For modality $m \in {R,T,E}$, the sampling process for the number of sampled point, $K$, yields the following aggregated features.
For modality $m \in {R,T,E}$, the sampling process for the number of sampled point, $Q  $, yields the following aggregated features.
\begin{equation}
% \mathbf{f}^{m,k}_j = \mathbf{F}_I^m(u_j^m + \Delta u^{m,k}_j),
\hat{\mathbf{f}}^m_j = \sum_{q=1}^Q w_q \cdot \mathbf{F}_I^m(u_j^m + \Delta u^{m,q}_j),
\end{equation}
where $\Delta u^{m,q}_j$ and $w_q$ are the learnable offset and aggregation weight for the $q$-th sampling point. Both the offsets $\Delta u^{m,q}_j$ and the weights $w_q$ are predicted from the voxel feature, $\mathbf{f}_{V_j}$. Each voxel feature $\mathbf{f}_{V_j}$ is treated as the query, while the aggregated image features $\hat{\mathbf{f}}^{m}_j$ serve as keys and values. The deformable cross-attention~\cite{vaswani2017attention} for voxel $V_j$ is formulated as
% \begin{equation}
% \hat{\mathbf{f}}^m_{V_j} =   \textit{Cross-Attn}\hat{\mathbf{f}}^{m}_j,
% \end{equation}
% \begin{equation}
$
\hat{\mathbf{f}}^m_{V_j} = \text{Attn}(\mathbf{Q} = \mathbf{f}_{V_j}, ~ \mathbf{K} = \hat{\mathbf{f}}^m_j ,~ \mathbf{V} = \hat{\mathbf{f}}^m_j).
$
% \label{eq:attention}
% \end{equation}

\setlength{\aboverulesep}{-1.5pt}
\setlength{\belowrulesep}{0pt}
\setlength{\tabcolsep}{7.6pt}
\renewcommand{\arraystretch}{1.05}
\begin{table*}[t]
\begin{center}
\caption{3D object detection performance comparison on the DSERT-RoLL dataset. We categorize the methods into three groups: stereo-based, 3D range sensor–based, and multi-modal fusion-based approaches.  For the modalities, we use the following notation: R: RGB, E: Event, T: Thermal, 4R: 4D Radar, and L: LiDAR. 
}
\vspace{-7pt}
\label{tab:main_3dod}
\resizebox{.99\linewidth}{!}{
\begin{tabular}{c|c|cccccc|cccc}
\toprule
\multirow{3}{*}{Modality} & \multirow{3}{*}{Methods} & \multicolumn{6}{c|}{Weather Condition}  & 
\multicolumn{4}{c}{Light Condition} \\
\cline{3-12}
&  &  \multirow{2}{*}{Clear} & \multirow{2}{*}{Fog} & Light & Heavy & Light & Heavy  & \multirow{2}{*}{Normal} & Low  & Over  & \multirow{2}{*}{HDR} \\
& & & & Rain & Rain & Snow & Snow & & Light & Expose & 
\\
\hline
\rowcolor{LG}
\multicolumn{12}{c}{\textbf{Stereo-based}} \\
\hline
R & DSGN~\cite{chen2020dsgn} & 31.08 & 43.66 & 42.48 & 20.51 & 25.94 & 0.01 & 29.99 & 25.68 & 22.55 & 40.69\\
R & LIGA~\cite{guo2021liga} & 35.52 & 41.67 & 37.52 & 20.57 & 26.02 & 0.00 & 31.31 & 30.06 & 22.44 & 42.80\\
E & DSGN~\cite{chen2020dsgn}&  24.23 & 22.06 & 26.93 & 31.38 & 23.12 & 0.01 & 21.41 & 21.42 & 15.58 & 36.44\\
E & LIGA~\cite{guo2021liga} &  27.11 & 22.53 & 23.43 & 22.84 & 24.61 & 0.00 & 23.10 & 23.20 & 15.30 & 34.92\\
T & DSGN~\cite{chen2020dsgn}& 28.49 & 25.98 & 37.50 & 28.74 & 36.52 & 0.02 & 16.89 & 36.07 & 25.83 & 36.03 \\
T & LIGA~\cite{guo2021liga} & 28.96 & 31.87 & 36.87 & 25.72 & 39.83 & 0.00 & 17.02 & 34.62 & 23.28 & 40.50\\
\hline
\rowcolor{LG}
\multicolumn{12}{c}{\textbf{3D Range Sensor-based}} \\
\hline
L & VoxelNeXt~\cite{chen2023voxelnext} & 86.06 & 59.51 & 90.19 & 71.82 & 82.86 & 54.75 & 78.93 & 88.76 & 71.06 & 80.93 \\
L & HEDNet~\cite{zhang2024hednet} & 79.27 & 48.41 & 84.74 & 68.36 & 70.29 & 55.98 & 71.64 & 83.34 & 63.97 & 73.33 \\
L & SAFDNet~\cite{Zhang_2024_CVPR} & 79.30 & 43.83 & 82.82 & 57.33 & 65.07 & 49.30 & 66.28 & 81.62 & 58.38 & 76.19 \\
4R & RTNH~\cite{paek2022k} & 23.49 & 37.30 & 43.40 & 27.86 & 36.96 & 21.70 & 28.70 & 26.28 & 24.69 & 27.00 \\
4R & VoxelNeXt~\cite{chen2023voxelnext} & 25.03 & 44.03 & 48.78 & 27.91 & 37.42 & 32.79 & 31.82 & 24.02 & 32.50 & 35.03 \\
4R & HEDNet~\cite{zhang2024hednet} & 24.10 & 43.51 & 41.16 & 28.57 & 31.28 & 25.67 & 28.92 & 22.28 & 37.01 & 30.82 \\
\hline
\rowcolor{LG}
\multicolumn{12}{c}{\textbf{Multi-modal Fusion-based}} \\
\hline
R+L & LoGoNet~\cite{li2023logonet} & 87.18 & 64.96 & 91.41 & 79.12 & 79.74 & 66.20 & 79.01 & 90.56 & 80.49 & 82.78 \\
R+L & BEVFusion~\cite{liu2022bevfusion} & 85.20 & 62.40 & 90.91 & 73.30 & 75.22 & 57.61 & 76.86 & 87.55 & 78.07 & 78.90 \\
R+L & DeepFusion~\cite{li2022deepfusion} & 87.19 & 63.94 & 91.91 & 75.61 & 81.77 & 57.26 & 79.19 & 90.81 & 78.66 & 80.10 \\
R+4R & HGSFusion~\cite{gu2025hgsfusion} & 25.74 & 46.49 & 49.62 & 28.49 & 37.87 & 34.02 & 32.66 & 24.31 & 34.47 & 35.96 \\
4R+L & InterFusion~\cite{wang2022interfusion} & 84.52 & 66.94 & 94.31 & 76.56 & 74.13 & 64.82 & 79.31 & 87.49 & 75.55 & 79.95 \\
4R+L & RL3DOD~\cite{chae2024towards} & 85.05 & 63.15 & 88.39 & 76.17 & 81.41 & 65.87 & 77.77 & 87.26 & 78.32 & 81.50\\
R+T+4R+L & SAMFusion~\cite{palladin2024samfusion} & 87.03 & 65.13 & 91.69 & 78.02 & 79.81 & 70.59 & 80.54 & 89.93 & 80.16 & 82.50\\
\rowcolor{bgrey}
R+E+T+4R+L & \textbf{Ours} & \textbf{90.30} & \textbf{71.42} & \textbf{95.10} & \textbf{80.26} & \textbf{85.59} & \textbf{72.94} & \textbf{82.93} & \textbf{92.65} & \textbf{85.47} & \textbf{86.33} \\
\bottomrule
\end{tabular}
}
\end{center}
\vspace{-18pt}
\end{table*}

\noindent
\textbf{Confidence-based Voxel Fusion.} 
We concatenate the image-enhanced voxel features from all modalities ($m\in\{R,T,E\}$) to form
\begin{equation}
\hat{\mathbf{F}}_V^{\text{cam}}
=
\big[\,\hat{\mathbf{F}}_V^{R}\ |\ \hat{\mathbf{F}}_V^{T}\ |\ \hat{\mathbf{F}}_V^{E}\,\big]
\in \mathbb{R}^{N_V \times (K C_V)},\quad K{=}3,
\end{equation}
where $\hat{\mathbf{F}}_V^{R}$, $\hat{\mathbf{F}}_V^{T}$, and $\hat{\mathbf{F}}_V^{E}$ denote the RGB, thermal, and event branches, respectively. For camera-axis attention, we view $\hat{\mathbf{F}}_V^{\text{cam}}$ as a 3D tensor, $\mathbb{R}^{N_V \times K \times C_V}.
$
We compute the camera-wise gates using a global summary as
\begin{equation}
\mathbf{w}
= \sigma\!\left(\frac{1}{N_V C_V}
\sum_{i=1}^{N_V}\sum_{c=1}^{C_V}
\hat{\mathbf{F}}_{V}^{\text{cam}}(i,:,c)\right)
\in [0,1]^{1 \times K \times 1},
\end{equation}
where $\sigma$ denotes the sigmoid activation. Each camera branch is reweighted by its corresponding scalar gate as:
\[
\bar{\mathbf{F}}_V^{cam}
= \mathbf{w} \odot \hat{\mathbf{F}}_V^{cam} 
\in \mathbb{R}^{N_V\times K\times C_V},
\]
where $\odot$ denotes element-wise multiplication. Finally, we concatenate the image-enhanced voxel features with the original voxel features.
A feed-forward network is then applied to reduce the channel dimension and yield the final fused features $\tilde{\mathbf{F}}_V \in \mathbb{R}^{N_V \times C_V}$.

\noindent
\textbf{Bounding Box Refinement.}
Given the initial bounding box proposals $\mathcal{B} = \{b_1, b_2, \ldots, b_n\}$ from the range sensors, we further perform refinement using the final fused voxel features $\tilde{\mathbf{F}}_V \in \mathbb{R}^{N_V \times C_V}$. 
For each proposal $b_i$, we divide the 3D region into $S \times S \times S$ regular sub-voxels and apply ROI pooling~\cite{hu2022point, deng2021voxel} to extract proposal-aligned features from both the fused image-enhanced voxel features $\tilde{\mathbf{F}}_V$ and the original voxel features. This produces grid features $\tilde{\mathbf{F}}_V^i \in \mathbb{R}^{S^3 \times C_V}$ for each initial bounding box, which are then passed through a multi-layer perceptron (MLP) to estimate the refined boxes, $\tilde{\mathcal{B}} = \{\tilde{b}_1, \tilde{b}_2, \ldots, \tilde{b}_n\}$.

\textbf{Loss Functions.} 
We train the entire framework in an end-to-end manner. The overall loss consists of three terms: the RPN loss~\cite{deng2021voxel, hu2022point} $\mathcal{L}_{\text{RPN}}$, the confidence prediction loss~\cite{deng2021voxel} $\mathcal{L}_{\text{conf}}$, and the box regression loss~\cite{deng2021voxel, shi2020pv} $\mathcal{L}_{\text{reg}}$:
\begin{equation}
\mathcal{L} = \mathcal{L}_{\text{RPN}} + \lambda_1\mathcal{L}_{\text{conf}} + \lambda_2 \mathcal{L}_{\text{reg}}.
\end{equation}
\section{Experiments on Multi-modal Approach}

\subsection{Experimental Settings}

We train the entire framework in an end-to-end manner using four NVIDIA Quadro RTX 8000 GPUs. The loss weights for both $\lambda_1$ and $\lambda_2$ are set to 1. For evaluation, to align with the front camera’s field of view, we constrain the point cloud along the X-axis to the range [0, 70] meters. We set the sampled points $K$ as 4 for feature aggregation in the camera-3D range sensor fusion module. Following prior work~\cite{deng2021voxel, hu2022point}, the bounding-box refinement grid size, $S$, is set to 6.
We evaluate all models using the official Waymo Open Dataset metrics~\cite{sun2020scalability}. We report Average Precision (AP) with a 3D IoU threshold of 0.5. Following prior work~\cite{paek2022k, bijelic2020seeing, kent2024msu} under the standard challenge conditions for 3D perception, our main tables emphasize the \textit{vehicle} class. The results for additional classes are provided in the supplementary material. We use a long-range LiDAR for the LiDAR modality, and for all camera modalities, we use only the left camera from the stereo setup.

\setlength{\belowrulesep}{0pt}
\setlength{\tabcolsep}{9.6pt}
\renewcommand{\arraystretch}{1.00}
\begin{table*}[t]
\begin{center}
\caption{2D object detection performance on the DSERT-RoLL dataset, focusing on camera-based methods. For the modalities, we use the following notation: R: RGB, E:
Event, and T: Thermal}
\vspace{-7pt}
\label{tab:main_2D}
\resizebox{.99\linewidth}{!}{
\begin{tabular}{c|c|cccccc|cccc}
\toprule
\multirow{3}{*}{Modality} & \multirow{3}{*}{Methods} &  \multicolumn{6}{c|}{Weather Condition}  & 
\multicolumn{4}{c}{Light Condition} \\ 
\cline{3-12}
 &  & \multirow{2}{*}{Clear} & \multirow{2}{*}{Fog} & Light & Heavy & Light & Heavy  & \multirow{2}{*}{Normal} & Low  & Over  & \multirow{2}{*}{HDR} \\
& & & & Rain & Rain & Snow & Snow & & Light & Expose & 
\\
\hline
R & YOLOv10~\cite{wang2024yolov10} & 76.47 & 72.99 & 84.95 & 76.68 & 58.76 & 2.84 & 71.98 & 67.69 & 76.25 & 76.15  \\
R & DEIM~\cite{huang2025deim} & 81.85 & 82.99 & 91.48 & 73.60 & 65.07 & 13.37 & 77.76 & 72.74 & 85.14 & 79.50 \\
E & RT-DETR~\cite{zhao2024detrs} & 73.77 & 83.17 & 83.57 & 58.93 & 47.28 & 0.023 & 69.89 & 58.28 & 78.87 & 77.83 \\
E & DEIM~\cite{huang2025deim} & 65.56 & 85.67 & 80.77 & 64.36 & 50.00 & 0.075 & 69.31 & 53.94 & 57.20 & 69.38 \\
T & YOLOv10~\cite{wang2024yolov10}& 78.31 & 83.84 & 92.16 & 76.75 & 75.30 & 0.619 & 69.48 & 74.15 & 73.93 & 81.03 \\
T & DEIM~\cite{huang2025deim} & 81.84 & 85.56 & 83.21 & 77.75 & 77.04 & 0.576 & 66.07 & 76.69 & 84.91 & 86.19 \\
R+E & GM-DETR~\cite{xiao2024gm} & 84.24 & 87.54 & 95.07 & 80.92 & 59.44 & 15.62 & \textbf{83.32} & 73.61 & 86.04 & 81.90\\
R+T & GM-DETR~\cite{xiao2024gm} & 84.10 & 86.64 & 92.18 & 77.99 & 79.44 & 1.48 & 71.87 & 77.41 & 86.12 & 88.70 \\
T+E & GM-DETR~\cite{xiao2024gm} & 85.44 & 92.13 & 88.35 & 79.64 & 81.19 & 11.20 & 71.00 & 78.96 & 87.74 & 93.04 \\
R+T+E & GM-DETR~\cite{xiao2024gm} & \textbf{90.36} & \textbf{93.66} & \textbf{96.28} & \textbf{82.29} & \textbf{81.60} & \textbf{16.56} & 82.07 & \textbf{82.60} & \textbf{94.93} & \textbf{93.52} \\
\bottomrule
\end{tabular}
}
\end{center}
\vspace{-20pt}
\end{table*}

\subsection{3D Object Detection Results Across Modalities}

The proposed method operates adaptively across diverse modality combinations, providing a framework that highlights the strengths of multi-modal fusion. To study these effects, we conduct ablations over different sensor combinations and report the results in Table~\ref{tab:main_modality}.

We begin with the most fundamental 3D range sensor, LiDAR, which serves as the base for initial bounding box estimation. From this foundation, we incrementally incorporate additional modalities to evaluate how each sensor contributes to detection robustness and accuracy under various environmental conditions. Adding RGB introduces richer semantics and contextual cues, improving category discrimination and boundary localization under moderate conditions. However, its impact is limited in extreme lighting and weathers, so the overall gains remain modest in the most challenging scenarios. Introducing 4D Radar further enhances spatial consistency and stability, especially in adverse weather (\eg~heavy snow) where LiDAR signals may degrade. The fusion of LiDAR and 4D Radar yields noticeable gain across most conditions, confirming the complementary nature of their geometric cues.
When event and thermal modalities are integrated, the model becomes more resilient to dynamic illumination changes and low-visibility environments. In particular, R+E+T+4R+L, which leverages all modalities, achieves the highest performance overall, demonstrating the framework’s ability to adaptively fuse heterogeneous inputs and fully exploit the advantages of multi-modal perception.

\section{Benchmarks on the DSERT-RoLL Dataset}

DSERT-RoLL enables fair, like-for-like evaluation across multiple modalities on the same scenes. With both 3D and 2D manual annotations and diverse weather and lighting conditions, it offers a rigorous testbed for robustness and generalization. 
% Beyond baselines, it supports transparent comparisons among camera-only, 3D range–sensor, and multi-modal fusion methods, revealing failure modes and trade-offs by modality.
Accordingly, we establish benchmarks for both 3D Object Detection and 2D Object Detection on DSERT-RoLL dataset.

\subsection{3D Object Detection Benchmark Results}

The selected 3D detector models, organized by sensor-configuration groups, are categorized into three types: stereo-based, 3D range sensor–based, and multi-modal fusion–based approaches.

\noindent
\textbf{Stereo camera-based methods.}
We select two existing methods, LIGA~\cite{guo2021liga} and DSGN~\cite{chen2020dsgn}. To isolate the effect of the sensor type, we keep the architectures and training settings identical and simply replace the camera input with event or thermal data.

\noindent
\textbf{3D range sensor-based methods.}
We select 3D range sensor–based methods, VoxelNeXt~\cite{chen2023voxelnext}, HEDNet~\cite{zhang2024hednet}, and SAFDNet~\cite{Zhang_2024_CVPR}, for our evaluation, and additionally adopted RTNH~\cite{paek2022k} for the 4D Radar modality. 
Since the LiDAR pipeline naturally extends to 4D Radar, these methods enable fair and consistent evaluation across 3D range-sensor modalities.

\noindent
\textbf{Multi-modal fusion methods.}
We include the following multi-modal methods as comparisons: LoGoNet~\cite{li2023logonet}, BEVFusion~\cite{liu2022bevfusion}, DeepFusion~\cite{li2022deepfusion}, HGSFusion~\cite{gu2025hgsfusion}, InterFusion~\cite{wang2022interfusion}, RL3DOD~\cite{chae2024towards}, and SAMFusion~\cite{palladin2024samfusion}.

As shown in Table~\ref{tab:main_3dod}, we observe that limited information settings, specifically stereo without explicit 3D depth and 4D radar only, tend to underperform, whereas LiDAR is generally strong thanks to accurate geometric cues. However, LiDAR performance drops in adverse weather such as fog and snow. In contrast, multi-modal methods compensate: cameras provide missing semantic detail and 4D radar adds weather robustness. Our approach adaptively fuses all available sensor types, delivering consistently strong results across weather and illumination conditions.

\subsection{2D Object Detection Benchmark Results}
Although this paper focuses on 3D detection, DSERT-RoLL also supports 2D detection research and provides a foundation for future multi-modal studies. To facilitate subsequent work, we establish a 2D benchmark and report baseline results using existing methods only. 
Specifically, we evaluate the camera-based methods YOLOv10~\cite{wang2024yolov10}, RT-DETR~\cite{zhao2024detrs}, and DEIM~\cite{huang2025deim} for the single-modality setting, and GM-DETR~\cite{xiao2024gm} for the multi-modal setting.
Consistent with our 3D evaluation, we report Average Precision (AP) at a 2D IoU threshold of 0.5 with an emphasis on the Vehicle category; results for additional categories are available in the supplementary material. For 2D detection, we use the left camera images for each modality. As shown in Table~\ref{tab:main_2D}, and in line with our 3D results, the multi-modal setting demonstrates greater robustness across diverse weather and illumination conditions.
\section{Conclusion}
We present DSERT-RoLL, a comprehensive multi-modal perception dataset featuring stereo Event–RGB–Thermal cameras, 4D radar, and dual LiDAR sensors. We establish benchmarks on DSERT-RoLL for both 3D and 2D object detection and introduce a modality-adaptive fusion baseline that strengthens detection under challenging weather and lighting conditions. We believe DSERT-RoLL will serve as a valuable foundation for future research, promoting progress in robust multi-modal perception and enabling more reliable 3D and 2D understanding under diverse real-world conditions.

\clearpage
\onecolumn
\section*{Supplementary Material}
\addcontentsline{toc}{section}{Supplementary Material}

This supplemental document provides additional information on the proposed dataset, \textbf{DSERT-RoLL}, additional details, results and comparisons.
\begin{itemize}
\item The license for the DSERT-RoLL dataset is provided in Section~\ref{sec:dataset_license};
\item More extensive comparisons with existing datasets, along with additional details, are provided in Section~\ref{sec:dataset_comp};
\item The criteria for distinguishing weather and lighting conditions are described in Section~\ref{sec:weater_light_criteria};
\item Data statistics of the proposed dataset are presented in Section~\ref{sec:ds};
\item Detailed information on the calibration between sensors of different modalities in Section~\ref{sec:calib};
\item Details on the annotation procedure and the provided annotations in Section~\ref{sec:anno};
\item Pixel-level alignment between cameras with different axes for the 2D detection is described in Section~\ref{sec:hogmoraphy};
\item The anonymization of sensitive personal information is described in Section~\ref{sec:prviacy};
\item The implementation details of the proposed multi-modal 3D detection model are provided in Section~\ref{sec:implementation_details};
\item Additional dataset samples are presented in Section~\ref{sec:dataset_samples};
\item Qualitative comparisons with other methods are provided in Section~\ref{sec:qualitative};
\item Experiment on sensitivity to extrinsic calibration errors in Section~\ref{sec:sensitivity};
\item Evaluation results across multiple classes are presented in Section~\ref{sec:quantitative};
\end{itemize}
\vspace{5pt}

\section{\hh{Dataset License}}
\label{sec:dataset_license}

The DSERT-RoLL dataset and accompanying code are provided strictly for research purposes and are distributed under the CC BY-NC 4.0 license, allowing use solely for non-commercial activities.

\section{\hh{Dataset Comparison}}
\label{sec:dataset_comp}

Table~\ref{tab:dataset_compare2} provides additional comparisons with other datasets that could not be fully covered in the main paper. Although many autonomous driving datasets have been introduced recently, to the best of our knowledge there is no dataset that simultaneously includes multiple novel sensors while also covering a wide range of weather and lighting conditions. This highlights and further emphasizes the unique advantages of our proposed dataset, which provides a unified benchmark for studying different sensor modalities across diverse environmental conditions.

\section{\hh{Weather and Lighting Conditions Descriptions}}
\label{sec:weater_light_criteria}

\vspace{-7pt}
\begin{table*}[h]
\caption{Detailed criteria for weather and light conditions.}
\vspace{-6pt}
\centering
\renewcommand{\tabcolsep}{16.0pt}
\renewcommand{\arraystretch}{1.1}
\resizebox{.99\textwidth}{!}{
\begin{tabular}{c|c|c}
\thickhline
Condition & Name & \multicolumn{1}{c}{Description and criteria}\\
\hline
\multirow{7}{*}{\thead{Weather\\Condition}} & Clear & Clear weather that does not meet the five weather conditions below. \\
& Fog & \thead{Weather conditions in which distant objects are dimly visible due to omnidirectional fog, \\ corresponding to regions and periods under officially issued weather advisories.} \\
& Light Rain & Weather conditions with a precipitation rate of up to 5 mm per hour. \\ 
& Heavy Rain & Weather conditions with a precipitation rate ranging from 10 to 15 mm per hour. \\ 
& Light Snow & Weather condition with a snowfall accumulation rate of less than 1 cm per hour.\\
& Heavy Snow & Weather condition with a snowfall accumulation rate of greater than 1 cm per hour.\\ 
\hline
\multirow{7}{*}{\thead{Light\\Condition}} 
& Normal & \thead{Standard lighting condition with balanced illumination \\ and no significant overexposure or underexposure.} \\
& Low Light & \thead{Condition with insufficient illumination, typically \\ resulting in reduced visibility and increased image noise.} \\
& Over Expose & Condition in which excessive brightness causes loss of detail in highlighted regions.\\ 
& HDR & \thead{Condition where multiple exposure levels are combined to capture \\ both dark and bright areas with enhanced contrast and detail.} \\
\thickhline
\end{tabular}
}
\vspace{-5pt}
\label{tab:detail_scene}
\end{table*}

The proposed DSERT-RoLL dataset includes a wide range of weather conditions, such as clear, fog, rain, and snow, as well as diverse light conditions, including normal, low-light, overexposed, and HDR scenarios. These variations allow the dataset to cover autonomous driving scenes across highly diverse environmental settings. All sequences are categorized according to the criteria we defined for each condition, and the dataset is organized following the detailed definitions presented in Table~\ref{tab:detail_scene}.

\section{Dataset Statistics}
\label{sec:ds}
\begin{figure*}[h]
\centering
\includegraphics[width=1.0\linewidth]{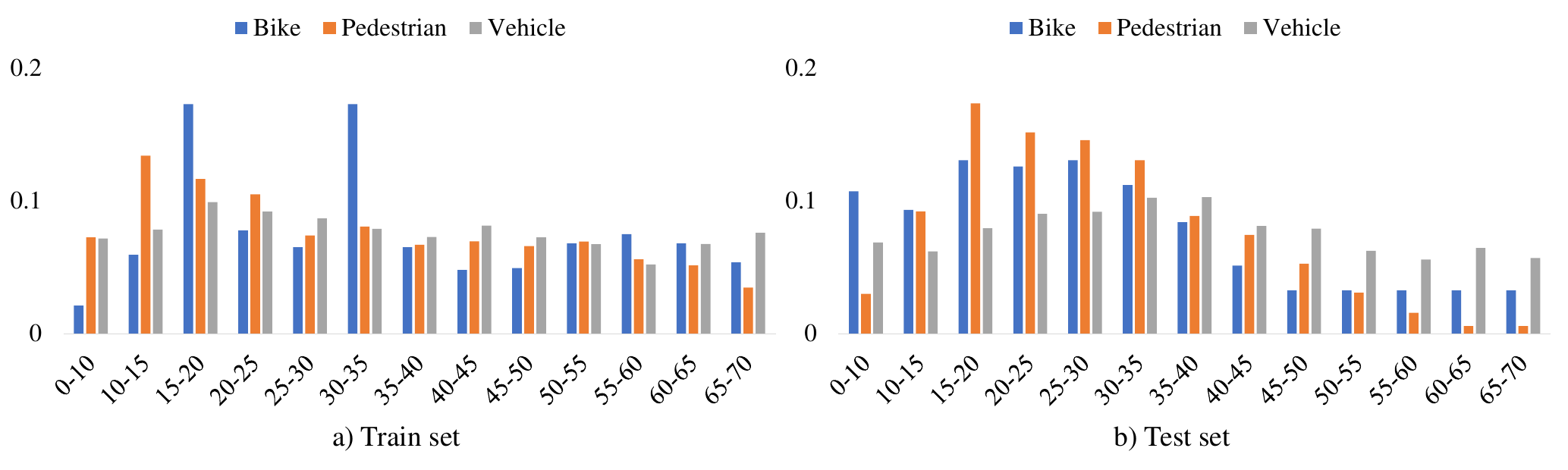}
    \vspace{-15pt}
    \caption{Class distribution across distance bins for the Bike, Pedestrian, and Vehicle categories. Panel a) shows the distribution in the training set, and panel b) shows the distribution in the test set.}
    \label{fig:class_distance}
    \vspace{-4pt}
\end{figure*}

\begin{figure*}[h]
\centering
\includegraphics[width=1.0\linewidth]{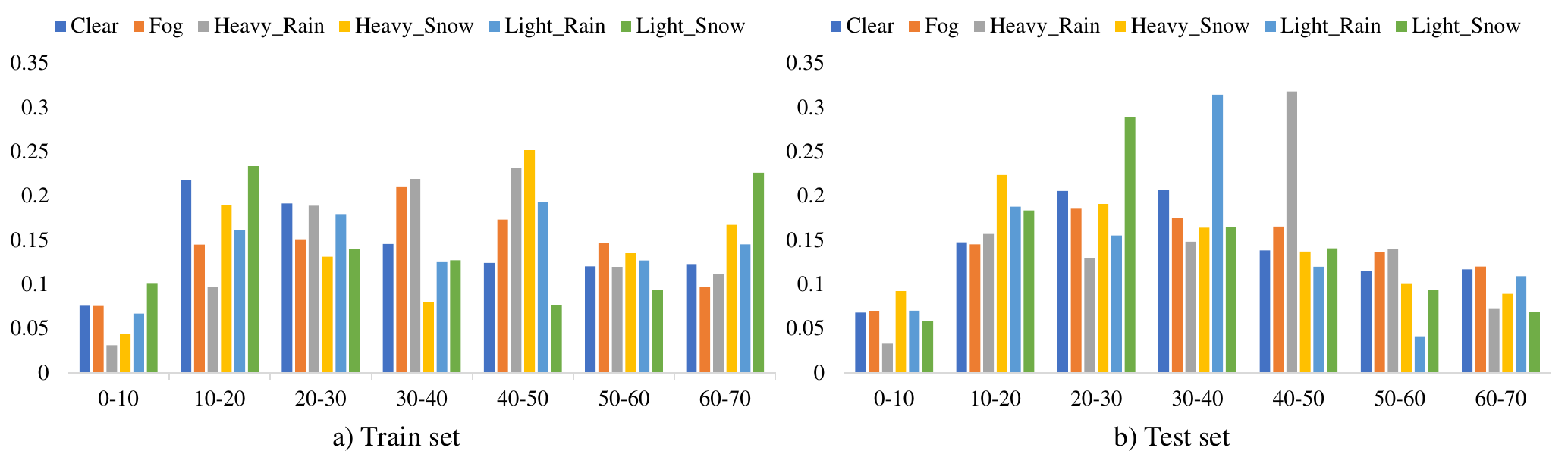}
    \vspace{-15pt}
    \caption{Weather condition distribution across distance bins. Each bar represents the proportion of Clear, Fog, Heavy Rain, Heavy Snow, Light Rain, and Light Snow in each distance interval. Panel a) shows the distribution in the training set, and panel b) shows the distribution in the test set.}
    \label{fig:weather_distance}
    \vspace{-4pt}
\end{figure*}

We analyze the distributions of object classes and weather conditions across distance intervals in both the training and test sets, as shown in~\cref{fig:class_distance,fig:weather_distance}. The results show broad coverage over the full distance range and similar tendencies between the two splits, indicating a limited distribution gap between training and evaluation.

As shown in~\cref{fig:class_distance}, the three object categories, Bike, Pedestrian, and Vehicle, are distributed across all distance bins in both splits. Although their proportions vary by interval, no class is concentrated within only a narrow distance range, which suggests balanced coverage over near, middle, and far regions.

A similar pattern is observed for weather conditions in~\cref{fig:weather_distance}. Clear, Fog, Heavy Rain, Heavy Snow, Light Rain, and Light Snow are all represented across the analyzed distance bins, and the train and test sets exhibit comparable trends. This indicates that the dataset remains reasonably balanced with respect to both semantic categories and environmental conditions over distance. Overall, these statistics support fair training and reliable evaluation under diverse real-world scenarios.

\section{Details of calibration}
\label{sec:calib}

In a multi-modal sensor system, calibration is essential for fusing and jointly using measurements from different sensors. To make this process convenient and effective, we use the RGB cameras as the reference modality and perform pairwise calibration with the other sensors in the following order:
[Sec.~\ref{sec:rgb_eventcalib}] calibration between RGB and the event stereo cameras, [Sec.~\ref{sec:rgb_thermalcalib}] calibration between RGB and the thermal stereo cameras, [Sec.~\ref{sec:rgb_lidar}] calibration between RGB and the LiDAR, and [Sec.~\ref{sec:rgb_4dradar}] calibration between RGB and the 4D radar.

\subsection{\hh{Calibration of the RGB and Event Stereo Cameras}}
\label{sec:rgb_eventcalib}

\begin{figure*}[h]
\centering
\includegraphics[width=1.0\linewidth]{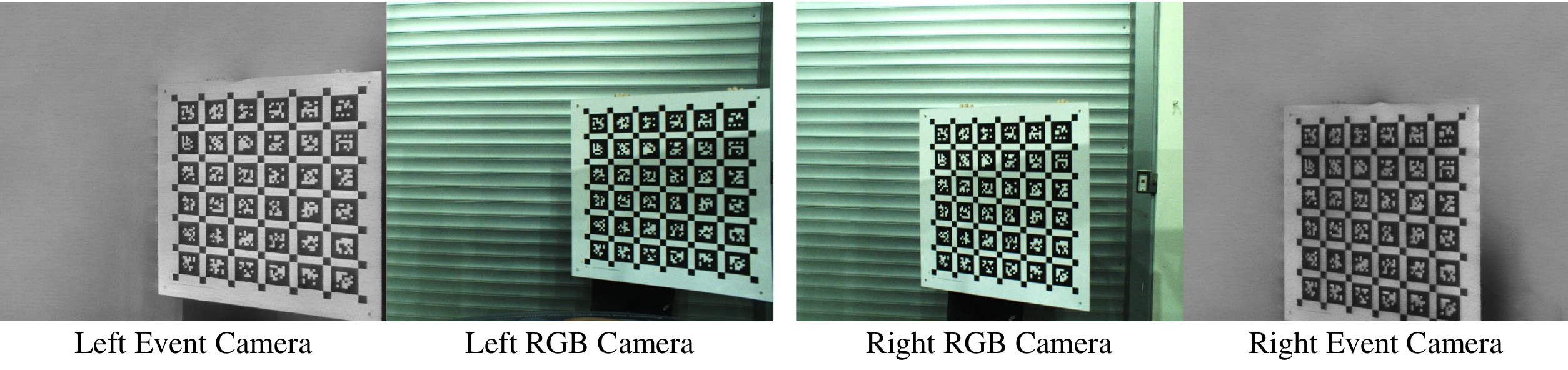}
    \vspace{-15pt}
    \caption{Sample data used for calibration, captured simultaneously by all four cameras observing the calibration pattern. The event data were reconstructed into intensity images using previous method~\cite{rebecq2019high}.
}
    \label{fig:rgb_event_calib}
    \vspace{-4pt}
\end{figure*}

To robustly calibrate the event cameras, we first collect a large amount of data from diverse viewpoints. Following prior work~\cite{Muglikar2021CVPR, gehrig2021dsec}, we employ an event-to-image reconstruction model~\cite{rebecq2019high} to convert the raw event streams into dense intensity images. As shown in Fig.~\ref{fig:rgb_event_calib}, this yields four synchronized image pairs from two RGB cameras and two event cameras, from which we construct multiple image pairs for calibration.
Using the widely adopted calibration toolbox Kalibr~\cite{furgale2013unified_kalibr}, we estimate the intrinsic parameters of each of the four cameras and jointly optimize the extrinsic parameters between all camera pairs in a single optimization step. This procedure allows us to obtain accurate intrinsic and extrinsic calibration for the four cameras.

\subsection{\yh{Calibration of the RGB and Thermal Stereo Cameras}}
\label{sec:rgb_thermalcalib}

\begin{figure*}[h]
\centering
\includegraphics[width=0.98\linewidth]{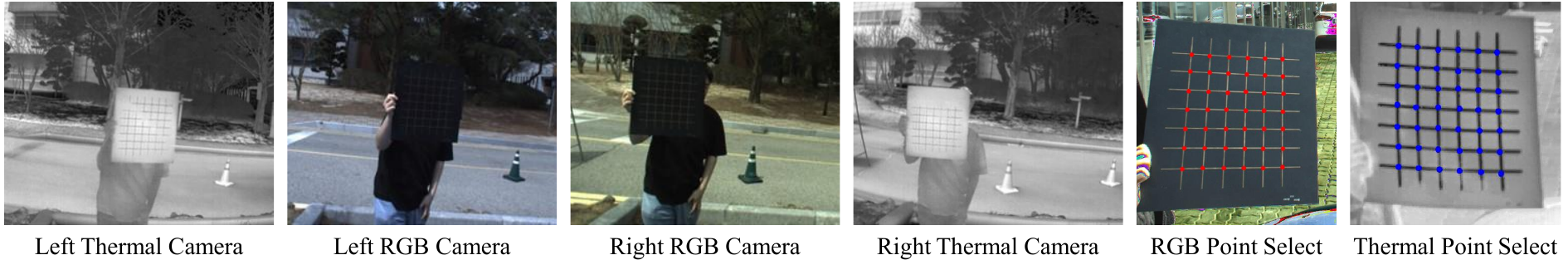}
    \vspace{-5pt}
    \caption{
    Sample data used for calibration, captured simultaneously by all four cameras observing the calibration pattern. We place points sequentially at the grid intersections and use them to construct the corresponding RGB-thermal point pairs.
}
    \label{fig:rgb_thermal_calib}
    \vspace{-4pt}
\end{figure*}

Because the standard calibration pattern is hardly visible in the thermal images, we resort to a manual RGB--thermal calibration procedure. To this end, we design a novel calibration target in which a grid of copper wires is mounted on a board so that the wires maintain a different temperature from the background surface and can be clearly captured by the thermal cameras. As shown in Fig.~\ref{fig:rgb_thermal_calib}, this setup yields four synchronized images from two RGB cameras and two thermal cameras, from which we construct multiple image pairs for calibration. We then place points sequentially at the grid intersections and use them to construct the corresponding RGB--thermal point pairs.

% \yh{RGB-Thermal}

\subsection{\hh{Calibration between RGB camera and LiDAR}}
\label{sec:rgb_lidar}

\begin{figure*}[h]
\centering
\includegraphics[width=0.86\linewidth]{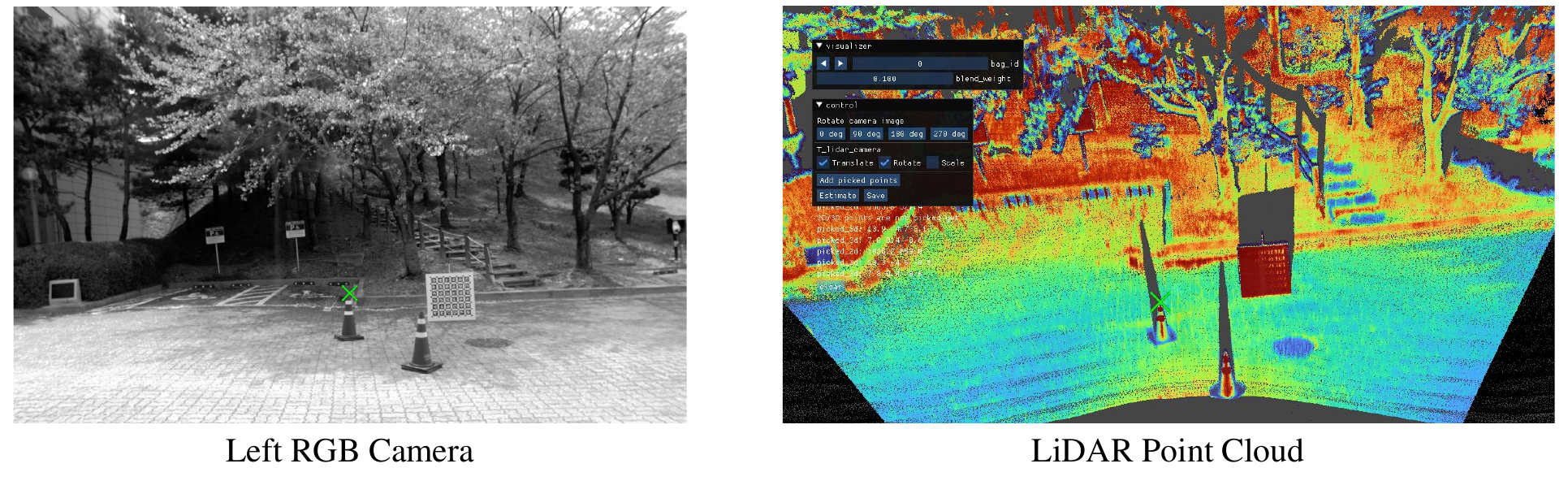}
    \vspace{-5pt}
    \caption{
    Example of a scene used by the calibration toolbox~\cite{koide2023general} for computing the extrinsic parameters between the RGB camera and the LiDAR.
}
    \label{fig:rgb_lidar_calib}
    \vspace{-4pt}
\end{figure*}

We use a recent LiDAR-camera extrinsic calibration toolbox~\cite{koide2023general} to estimate accurate transformations between each LiDAR and the reference RGB camera. As shown in Fig.~\ref{fig:rgb_lidar_calib}, the toolbox first constructs a dense LiDAR point cloud and derives an initial alignment with the camera using geometric correspondences. It then refines the extrinsic parameters through error-based optimization, yielding a precise calibration. The same procedure is applied to both the long-range and short-range LiDARs in our setup.

% \yh{RGB-4D Radar}

\subsection{\yh{Calibration between RGB camera and 4D Radar}}
\label{sec:rgb_4dradar}
\begin{figure*}[h]
\centering
\includegraphics[width=0.98\linewidth]{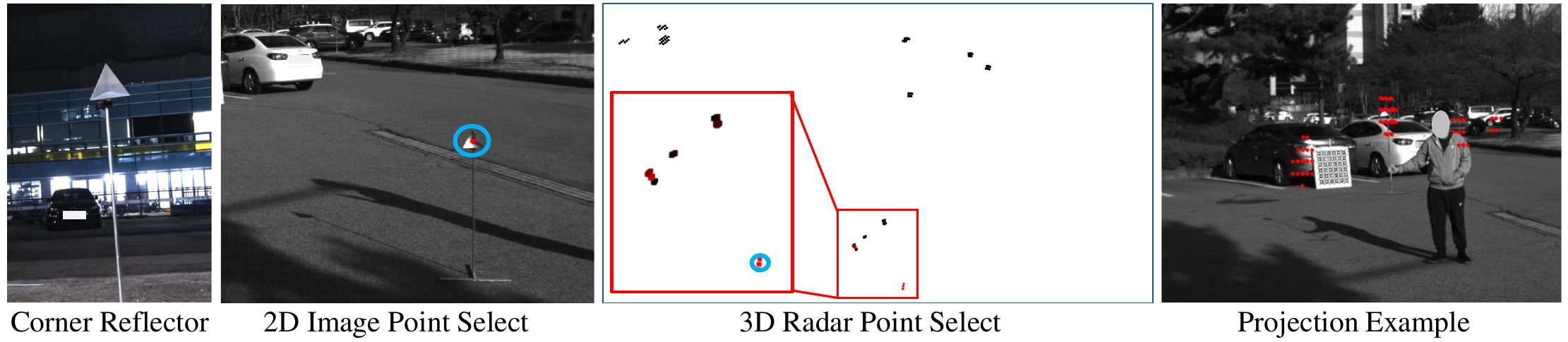}
    \vspace{-5pt}
    \caption{
    Example scene used for RGB–4D Radar calibration. Using the 4D Radar reflector target, we obtain 2D–3D point correspondences at various distances.
}
    \label{fig:rgb_radar_calib}
    \vspace{-4pt}
\end{figure*}

We calibrate the left RGB camera and radar by placing a corner reflector target in front of the sensor rig, causing the radar returns to collapse into a single high-intensity point. We record radar point clouds with 3D location, power, and Doppler attributes together with time-synchronized RGB images. Using an in-house annotation tool, we manually click the reflector in the radar point cloud and the corresponding pixel in the RGB image for each synchronized pair, yielding a set of 3D–2D correspondences.
Since the reflector concentrates the incident energy, the radar return power becomes significantly higher. We therefore select, in the 3D radar point cloud, the point with the highest return power as its 3D counterpart, as shown in~\cref{fig:rgb_radar_calib} at the location indicated by the light-blue circle.
From these correspondences, we solve for the rigid transformation between the radar and camera coordinate systems, enabling accurate projection of radar point clouds onto the image plane, while the remaining sensing modalities are tied to the radar via their pre-calibrated extrinsic parameters.

\subsection{\yh{Stereo Camera Rectification}}

\begin{figure*}[h]
\centering
\includegraphics[width=1.0\linewidth]{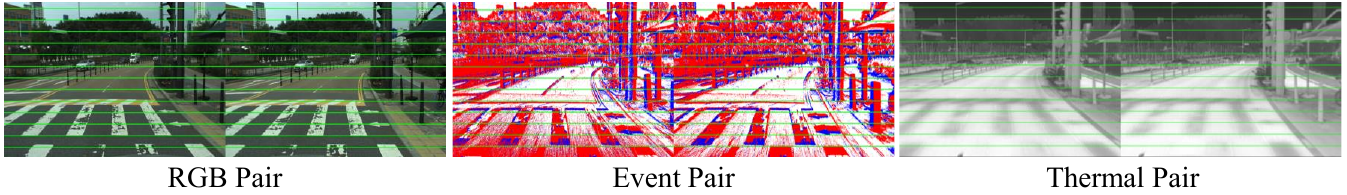}
    \vspace{-15pt}
    \caption{
    Example of stereo rectification results for the RGB, event, and thermal modalities. The green horizontal lines visualize the epipolar lines, which become well aligned after calibration. 
}
    \label{fig:stereo}
    \vspace{-4pt}
\end{figure*}

The proposed DSERT-RoLL dataset provides stereo data for all camera modalities.
For stereo rectification, we process each modality (RGB, event, and thermal) independently. Given the intrinsic matrices $K_L, K_R$, the distortion coefficients, and the relative pose $(R, t)$ obtained from our calibration, we compute the rectifying rotations $R_L, R_R$ and the new projection matrices $P_L, P_R$ using OpenCV's \texttt{stereoRectify}. We then undistort and warp the left and right images with \texttt{initUndistortRectifyMap} and \texttt{remap}, so that corresponding points in each stereo pair lie on the same scanline and the epipolar lines become approximately horizontal. The same procedure is applied to the RGB, event, and thermal stereo pairs. We provide sample stereo-rectified images in Fig.~\ref{fig:stereo}, where the rectification quality can be visually inspected. As shown in the figure, corresponding points between the left and right cameras of the same modality lie on the same horizontal scanline, confirming proper stereo rectification.

% \yh{Calibration Overalp Viz}

\subsection{\yh{Calibration Results of All Sensors}}
\begin{figure*}[p]
    \centering
    \includegraphics[width=0.87\linewidth]{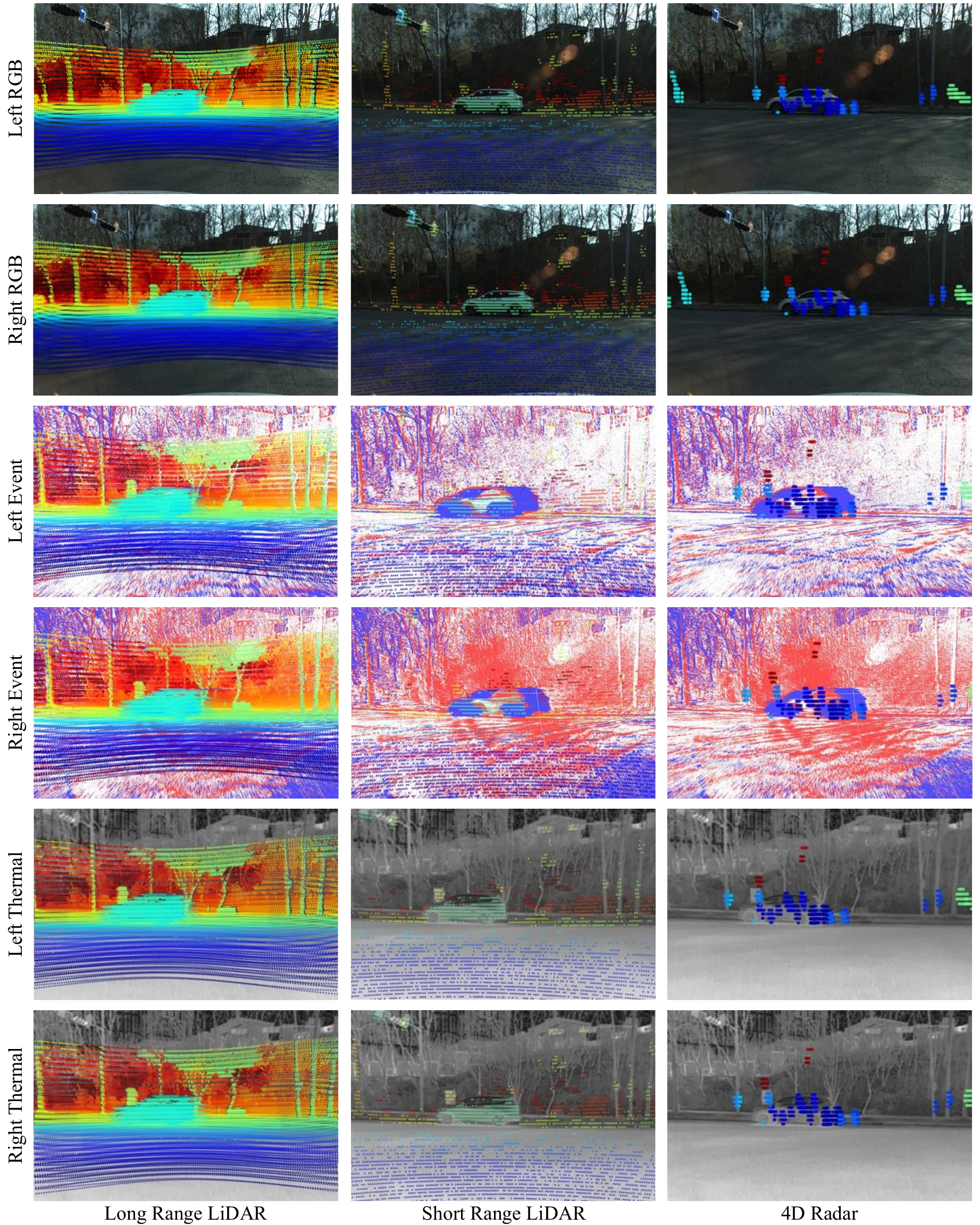}
    % \vspace{-18pt}
    \caption{Calibration results for all sensors, accompanied by an example scene illustrating the projection of 3D range sensor measurements onto the image planes of all six modalities.}
    \label{fig:all}
    % \vspace{-3pt}
\end{figure*}

Through the calibration steps described above, we obtain a consistent extrinsic calibration that links all sensing modalities. \Cref{fig:all} shows a representative scene demonstrating the quality of this calibration: 3D points from the range sensors are projected onto the image planes of the six camera modalities (left and right images of RGB, event, and thermal), demonstrating good spatial alignment across all views.

\section{\jy{Details of Annotation}}
\label{sec:anno}

\subsection{Annotation Procedure}

\begin{figure*}[h]
\centering
\includegraphics[width=0.8\linewidth]{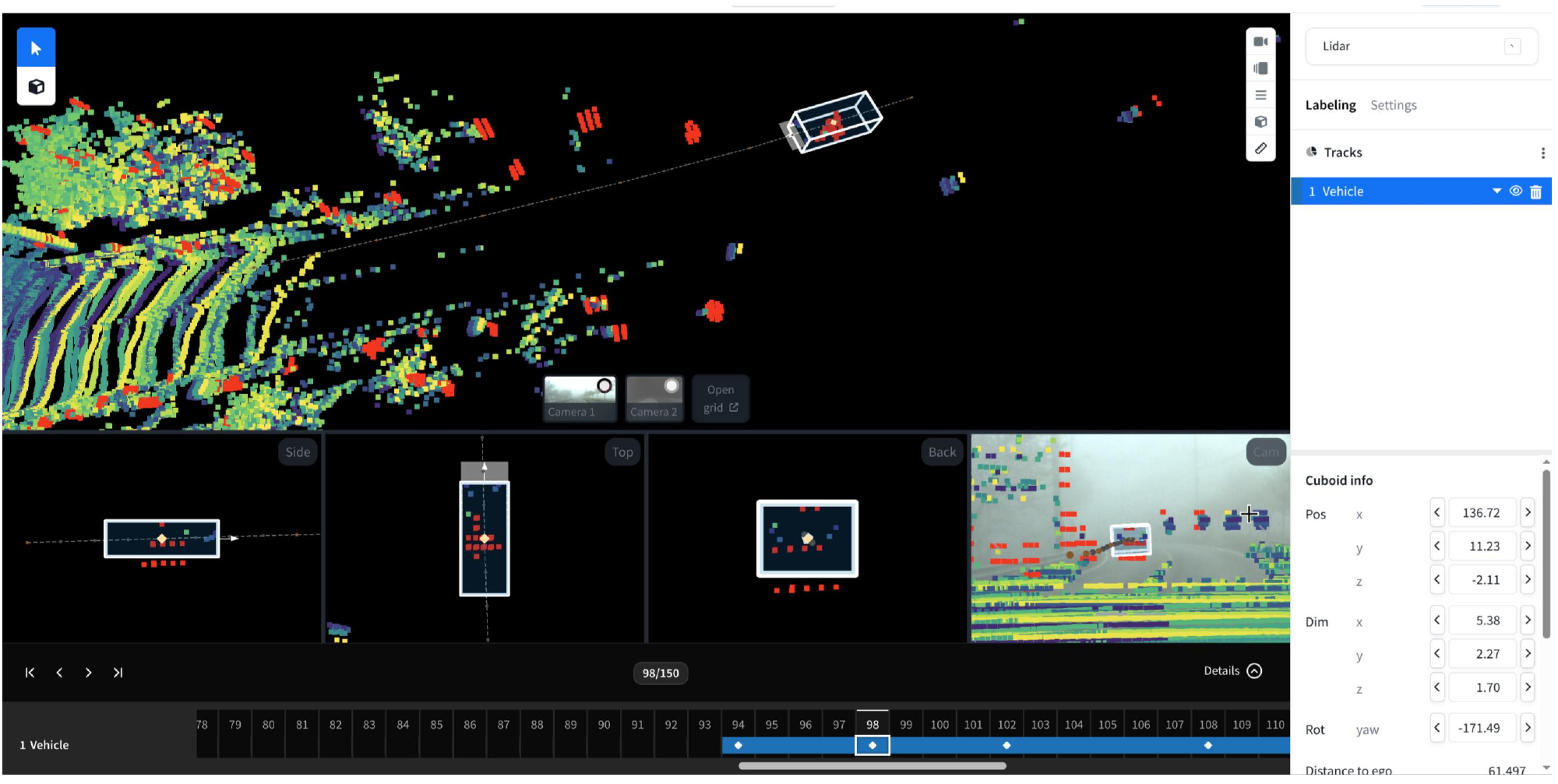}
    \vspace{-5pt}
    \caption{Multi-modal annotation tool. 4D Radar points are shown in red, and LiDAR points are color-coded by timestamp using a viridis colormap. 3D sensor data and 3D annotations can be projected onto camera sensor images, ensuring consistent annotations across modalities under diverse driving conditions..
}
    \label{fig:anno_tool}
    \vspace{-4pt}
\end{figure*}

As shown in Fig.~\ref{fig:anno_tool}, to obtain reliable annotations under diverse adverse weather and lighting conditions, we jointly exploited multiple sensor modalities during the labeling process. Specifically, we used two complementary 3D range sensors, LiDAR and 4D Radar, to provide accurate geometric information, while RGB and thermal images were employed to verify the alignment and validity of the bounding boxes. This multi-modal setup improves the accuracy of the ground-truth bounding boxes and helps ensure that no objects are missed during annotation. 

All multi-modal data were imported into a dedicated professional annotation tool~\cite{segmentsai}, and labeling was carried out by hired expert annotators. Fig.~\ref{fig:anno_tool} illustrates the annotation tool interface used by annotators to label multi-modal sensor data. In the main view, LiDAR and RADAR points are plotted in different colors, providing reliable 3D information even under adverse weather conditions (\eg, snow, fog). The 3D sensor data and 3D bounding boxes are projected onto the images, assisting annotators during labeling and enabling consistent annotations across modalities. Each video sequence was subsequently reviewed by at least three annotators to guarantee high-quality labels. 

For 2D bounding box annotation, 3D annotations are projected onto the image plane and subsequently refined. Odometry is defined as the relative pose with respect to the first sample of each video sequence and is estimated using an IMU–LiDAR-coupled SLAM algorithm~\cite{legoloam2018shan}, with GNSS signals used to aid pose estimation when 3D sensors are unreliable due to weather condition.

\subsection{Types of Annotations}
Table~\ref{tab:anno_type} summarizes the annotation types provided in the DSERT-RoLL dataset. Odometry is defined as the relative pose with respect to the first frame of each video sequence. The weather and lighting conditions are annotated once per sequence, meaning that a single sequence-level label is assigned rather than frame-level labels. The 3D annotations follow the Waymo Open Dataset (WOD)~\cite{sun2020scalability} format, while the 2D annotations are provided in the COCO~\cite{lin2014microsoft} format.

\begin{table*}[h]
\caption{Provided annotation types and descriptions.}
\vspace{-5pt}
\centering
\renewcommand{\tabcolsep}{12.0pt}
\renewcommand{\arraystretch}{1.1}
\resizebox{.75\textwidth}{!}{
\begin{tabular}{c|c|c}
% \toprule
\thickhline
Type & Format & \multicolumn{1}{c}{Description}\\
% \midrule
\hline
3D & 3D Bbox & $(center\_x, center\_y, center\_z, l, w, h, yaw, class )$\\
2D & 2D Bbox & $(u_{min},v_{min},w,h,class)$\\
Odometry & pose & $\mathbf{p}=[\mathbf{R} \mid \mathbf{t}]$ where $\mathbf{R} \in SO(3)$ and $\mathbf{t} \in \mathbb{R}^3$\\
Weather Condition & text & Sequence-level weather condition\\
Lighting Condition & text & Sequence-level lighting condition\\

\thickhline
\end{tabular}
}
\vspace{-10pt}
\label{tab:anno_type}
\end{table*}

\section{\yh{Image Homography for 2D Detection}}
\label{sec:hogmoraphy}

\begin{figure*}[h]
\centering
\includegraphics[width=0.93\linewidth]{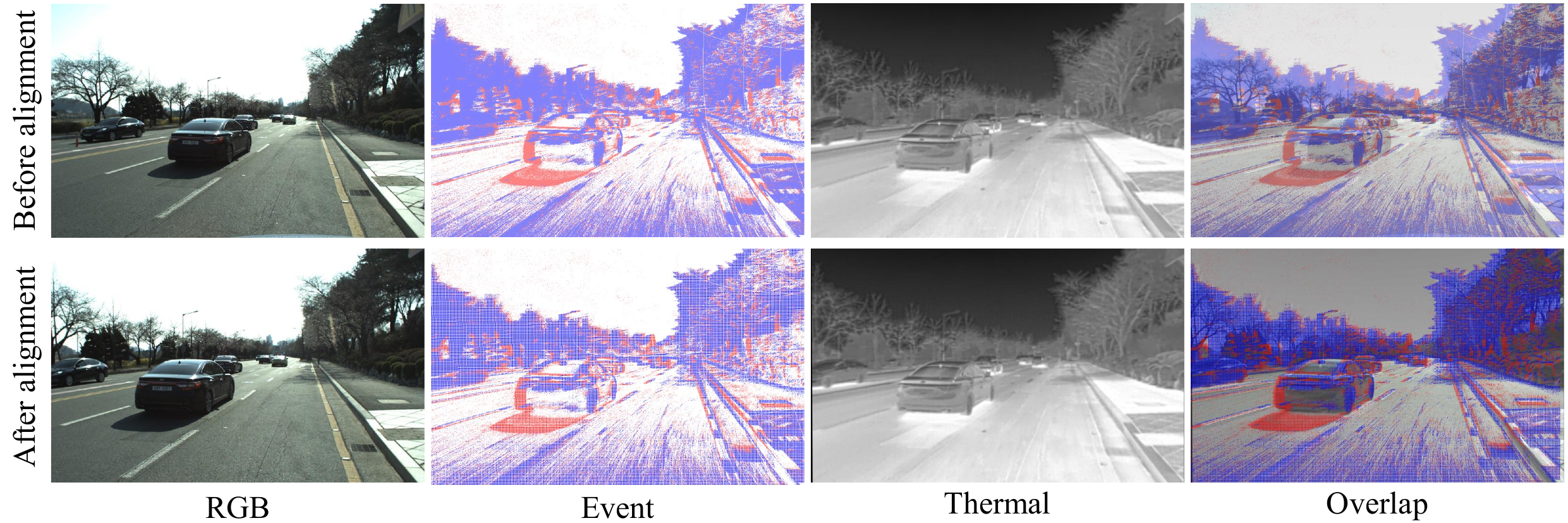}
    \vspace{-5pt}
    \caption{
    Example of RGB–event–thermal alignment using the homography. The top row shows the three modalities before alignment, and the bottom row shows the event and thermal images after being warped onto the RGB image plane and cropped to the common field of view, with the last column visualizing their overlap.
}
    \label{fig:homography}
    \vspace{-4pt}
\end{figure*}

The proposed multi-modal 3D object detection framework operates by projecting 3D information onto each camera and sampling in the image space, and therefore does not require the images themselves to be mutually aligned. However, for fair comparison in 2D object detection and to enable a pixel-level multi-modal fusion approach, pixel-wise alignment between the different image modalities is necessary.
To this end, we align thermal and event images to the RGB camera using a rotation-based image homography derived from the calibrated intrinsics and extrinsics of all sensors, following the prior work~\cite{gehrig2021dsec}. We do not perform any additional dedicated calibration for this alignment step; instead, we reuse the existing calibration parameters and treat the resulting homography as an approximate alignment. From the LiDAR-to-camera extrinsic matrices, we extract the $3 \times 3$ rotation matrices and compute the relative rotation between each source camera (thermal or event) and the RGB camera. After rescaling each camera's intrinsic matrix to the target resolution, we construct the homography using the infinite-plane approximation
\begin{equation}
    H = K_{\mathrm{RGB}}\, R_{\mathrm{RGB,src}}\, 
    K_{\mathrm{src}}^{-1},
\end{equation}
where each $K$ denotes the intrinsic parameters of the corresponding camera, and $R$ represents the extrinsic parameters that transform points from the source camera to the RGB camera coordinate system. We apply homography to warp the thermal and event images onto the RGB image plane, followed by cropping the overlapping field of view shared by all three modalities. An example scene is following~\cref{fig:homography}. The event image exhibits a grid-like appearance because the sparse events are forward-warped and dispersed over the RGB image plane.

\section{\hh{Privacy Concerns}}
\label{sec:prviacy}

\begin{figure*}[h]
    \centering
    \includegraphics[width=1.0\linewidth]{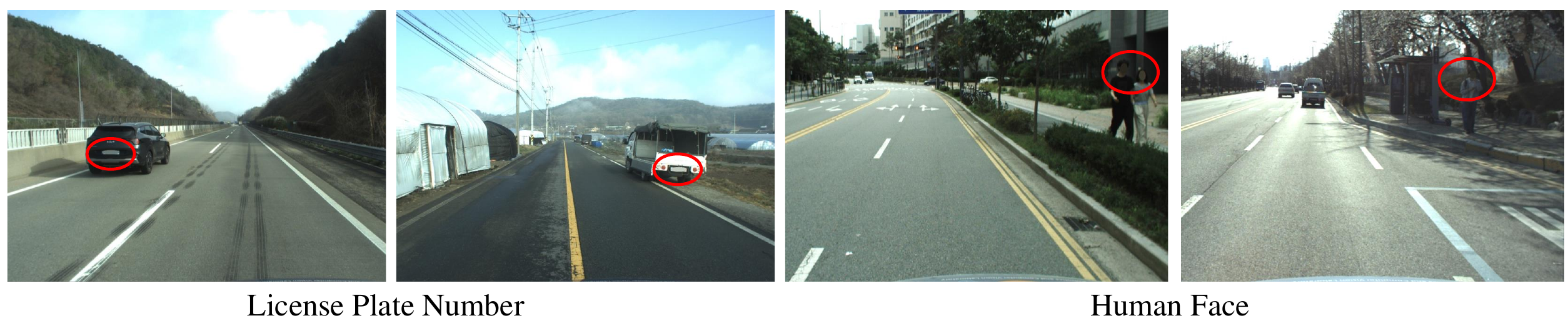}
    \vspace{-18pt}
    \caption{Example images in which privacy-sensitive information, such as license plate numbers and human faces, has been blurred.}
    \label{fig:privacy}
    \vspace{-3pt}
\end{figure*}

To address privacy concerns, we ensure that all privacy-sensitive information in the dataset is properly anonymized. In particular, all human faces and vehicle license plates are blurred to prevent any form of personal identification, as illustrated in Fig.~\ref{fig:privacy}.

\section{\hh{Implementation Details}}
\label{sec:implementation_details}

We train our multi-modal 3D object detection model in an end-to-end manner for a total of 20 epochs. The batch size is set to 8 per GPU, and all experiments are conducted using four NVIDIA Quadro RTX 8000 GPUs. We set the 3D detection range to $[0, 75.2\,\text{m}]$ along the $X$ axis, $[-75.2\,\text{m}, 75.2\,\text{m}]$ along the $Y$ axis, and $[-2\,\text{m}, 4\,\text{m}]$ along the $Z$ axis.
For each attention operation, we sample $Q = 4$ points (Eq.~(\cvtext{3}) in the main paper).
Following~\cite{deng2021voxel}, the grid size $S$ for the box refinement is set to $6$. 
For the each image branch, we use a Swin-Tiny~\cite{liu2021swin} backbone together with an FPN~\cite{lin2017feature}. We perform feature fusion using only the left cameras. The event data are converted into a tensor-based voxel grid representation~\cite{zhu2019unsupervised} with a bin size of 5, while the thermal data are used as single-channel images. To reduce computational cost, we downsample the RGB, thermal, and event inputs to $1/4$, $1$, and $1/2$ of their original resolution, respectively.
During training, we adopt common data augmentation techniques: random flips along the horizontal axis, global scaling with factors drawn from $[0.95, 1.05]$, and rotations about the $Z$ axis sampled from $[-\pi/4, \pi/4]$. After prediction, we apply non-maximum suppression (NMS) with an IoU threshold of $0.7$ to filter out overlapping detections and keep a single bounding box per object.

\section{\yh{More Dataset Samples}}
\label{sec:dataset_samples}

\begin{figure*}[p]
\centering
\includegraphics[width=1.0\linewidth]{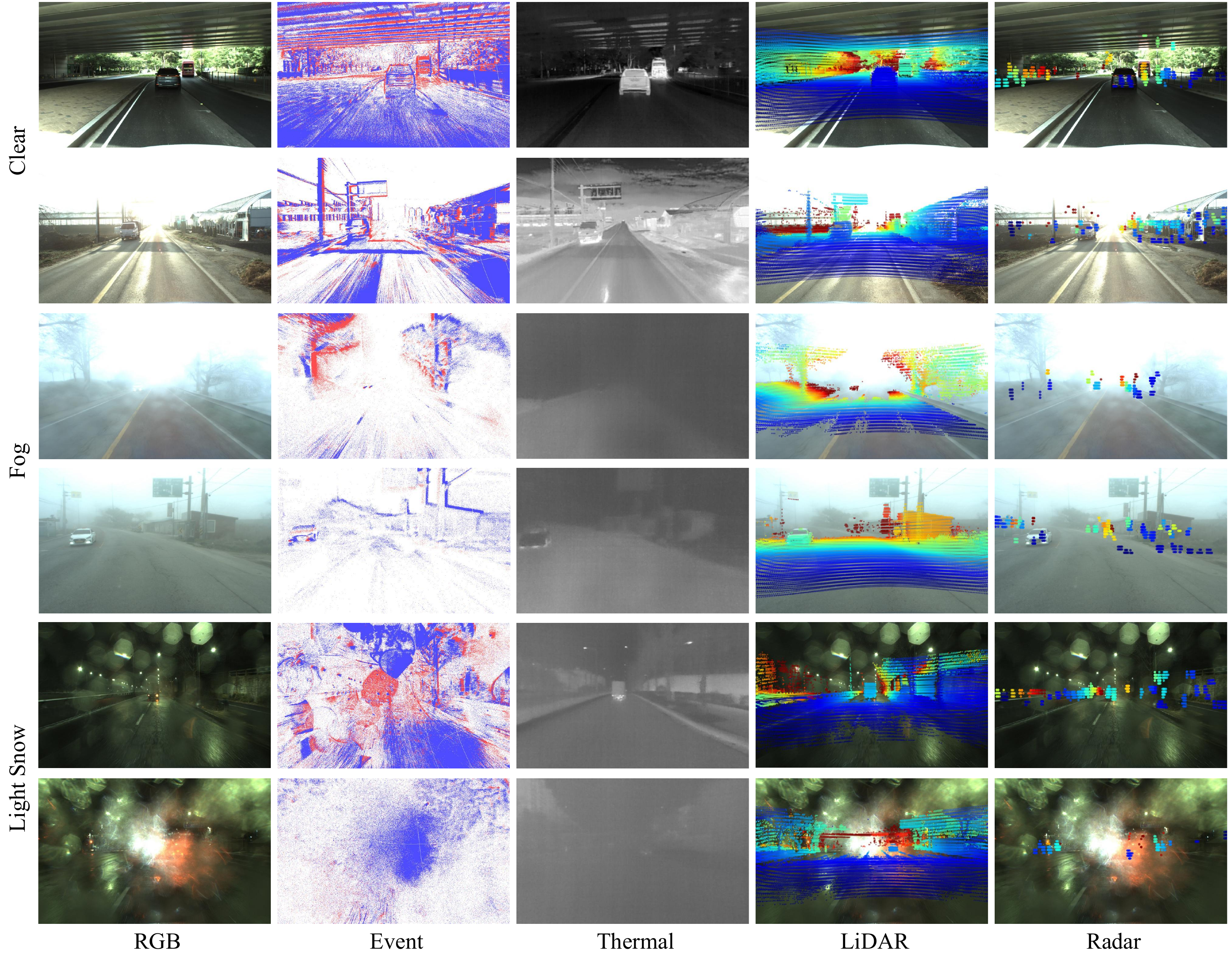}
    % \vspace{-15pt}
    \caption{More data samples for clear, fog and, light snow weather conditions. All 2D camera images are from the left camera. We project 3D range sensor data onto the left RGB camera image.}
    \label{fig:more_qual1}
    % \vspace{-4pt}
\end{figure*}

\begin{figure*}[p]
\centering
\includegraphics[width=1.0\linewidth]{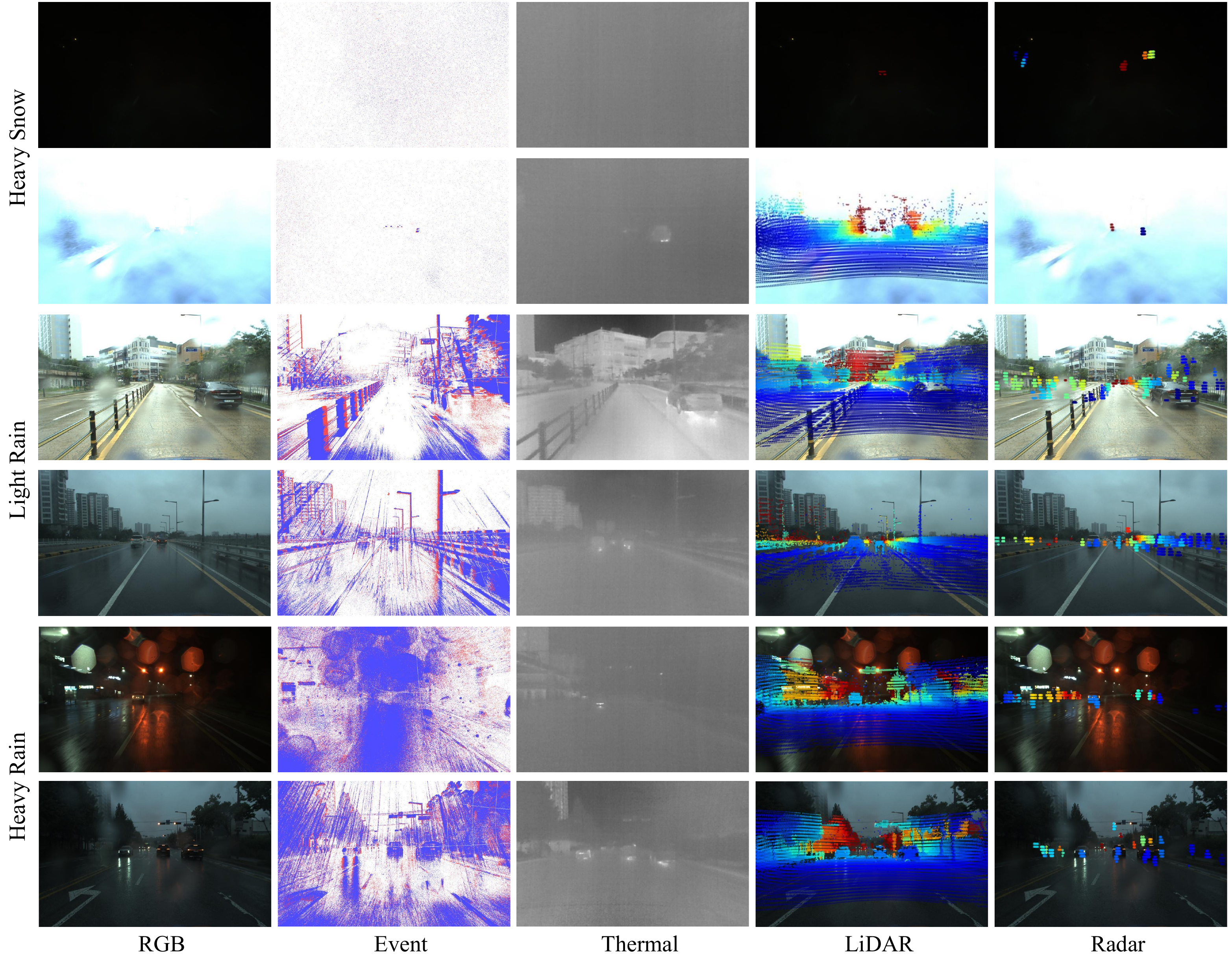}
    % \vspace{-15pt}
    \caption{More data samples for heavy snow, light rain and, heavy rain weather conditions. All 2D camera images are from the left camera. We project 3D range sensor data onto the left RGB camera image.}
    \label{fig:more_qual2}
    % \vspace{-4pt}
\end{figure*}

\begin{figure*}[p]
\centering
\includegraphics[width=1.0\linewidth]{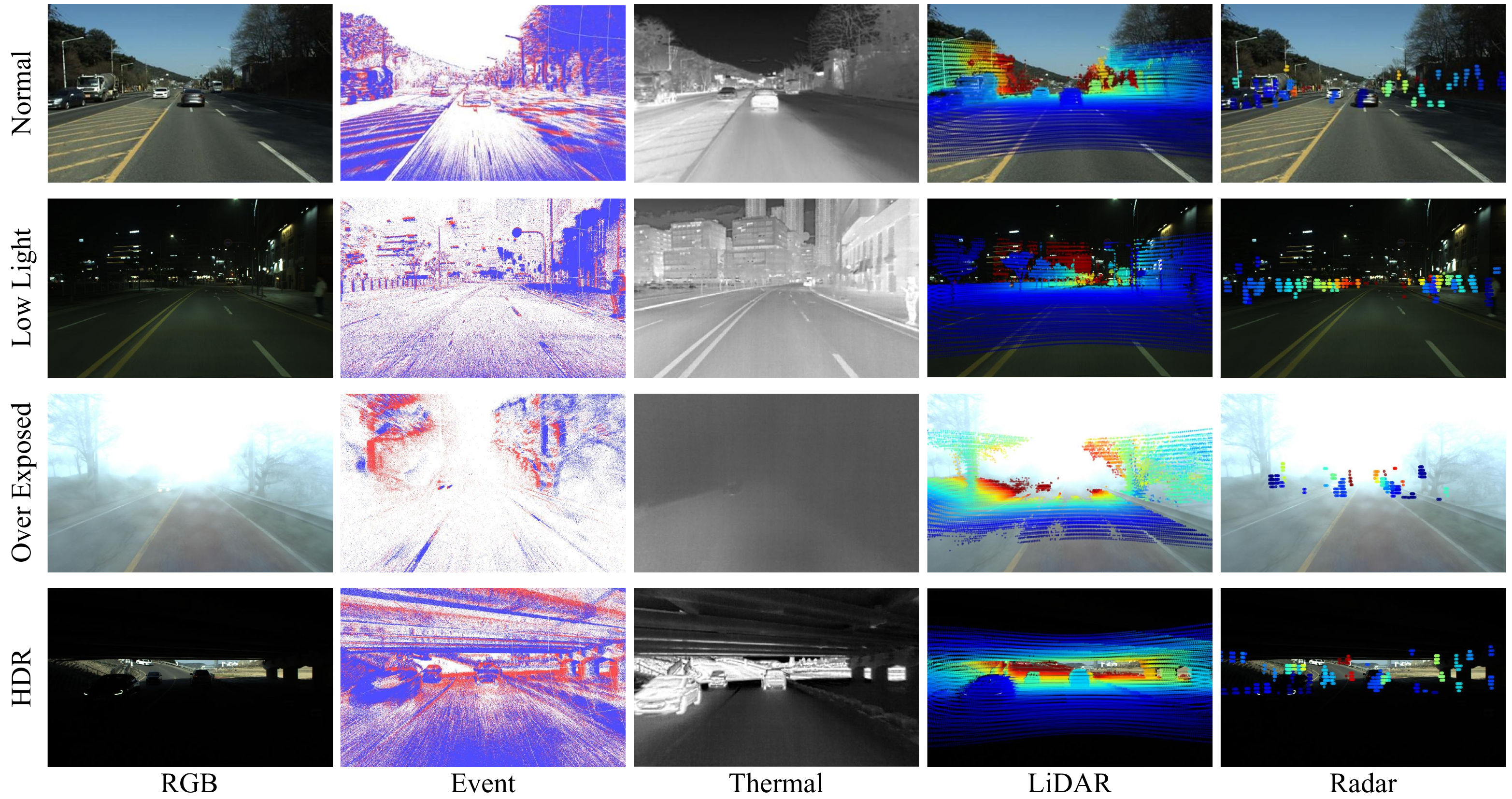}
    % \vspace{-15pt}
    \caption{More data samples for light conditions. All 2D camera images are from the left camera. We project 3D range sensor data onto the left RGB camera image.}
    \label{fig:more_qual_light}
    % \vspace{-4pt}
\end{figure*}

We provide additional dataset samples in~\cref{fig:more_qual1,fig:more_qual2} to illustrate the diversity of weather conditions. \Cref{fig:more_qual1} presents scenes under clear, fog, and light snow, whereas~\cref{fig:more_qual2} includes samples from heavy snow, light rain, and heavy rain. Moreover, we illustrate dataset samples of varying light conditions in~\cref{fig:more_qual_light}.

\section{Qualitative Results}
\label{sec:qualitative}
\begin{figure*}[p]
\centering
\includegraphics[width=0.99\linewidth]{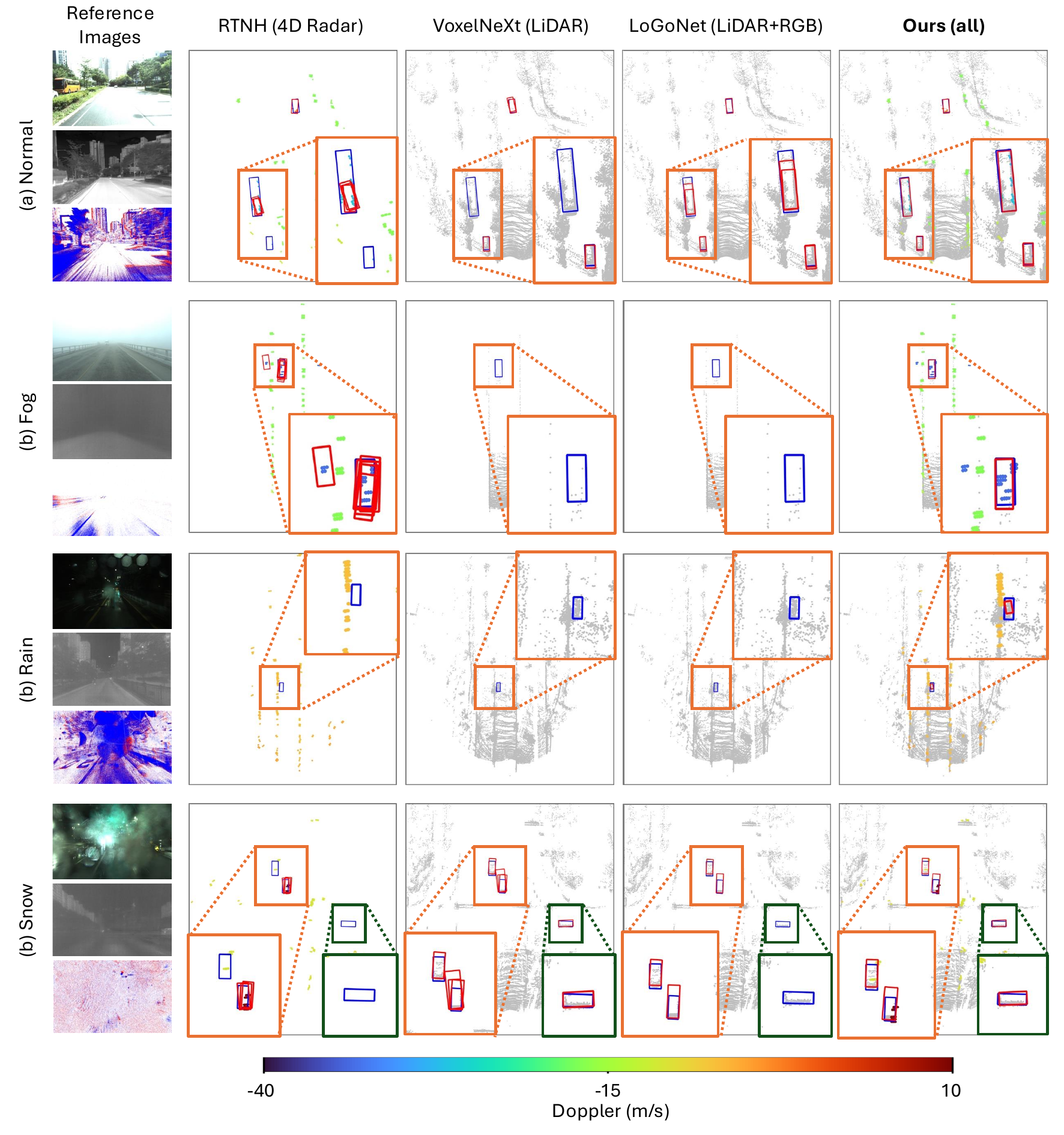}
    % \vspace{-15pt}
    \caption{Qualitative comparison of 3D detection results on the DSERT-RoLL dataset in the BEV (bird’s-eye-view) plane. 4D Radar-based (RTNH~\cite{paek2022k}), LiDAR-based (VoxelNeXt~\cite{chen2023voxelnext}), and LiDAR–RGB fusion (LoGoNet~\cite{li2023logonet}) methods are shown for comparison, and Ours uses all available sensors (RGB, event, thermal, 4D Radar, and LiDAR). Regions containing objects are highlighted and zoomed in for better visibility. 
    We visualize only the 3D range sensor used by each method. LiDAR measurements are shown as gray point clouds, while 4D radar points are colored according to their Doppler velocities. \textcolor{blue}{Blue boxes} denote the 3D ground truth, and \textcolor{red}{red boxes} denote the predicted 3D bounding boxes.
}
    \label{fig:supple_qual}
    \vspace{-4pt}
\end{figure*}

Fig.~\ref{fig:supple_qual} presents qualitative 3D detection results on the DSERT-RoLL dataset. For comparison, we include Radar-based RTNH~\cite{paek2022k}, LiDAR-based VoxelNeXt~\cite{chen2023voxelnext}, and LiDAR–RGB fusion LoGoNet~\cite{li2023logonet}. Our method leverages all available sensors (RGB, event, thermal, 4D Radar, and LiDAR), and the proposed multi-modal approach demonstrates robust performance across a wide range of weather conditions.

\noindent
\textbf{Normal. }Under normal conditions, all methods show reasonable performance and successfully detect distant objects as well as objects in the opposite lane. However, RTNH suffers from degraded detection performance when 4D Radar returns are sparse or only partially observed. VoxelNeXt also fails to detect the bus in this scenario. LoGoNet successfully detects all objects by exploiting sensor fusion, but the estimated object sizes are inaccurate. In contrast, our model accurately predicts both the positions and sizes of all objects.

\noindent
\textbf{Fog. }In foggy scenes, the range and quality of LiDAR measurements degrade significantly, causing VoxelNeXt and LoGoNet—both heavily dependent on LiDAR—to miss the object. RTNH, which relies on robust 4D Radar, detects the targets but produces several false positives due to sensor noise. Our model, on the other hand, leverages 4D Radar for object detection and is effectively guided by image sensors, resulting in more accurate and stable predictions.

\noindent
\textbf{Rain. }In the rainy-condition example, only our model successfully detects the bike. Radar is inherently weak at capturing small objects, and LiDAR performance also degrades under rain, which increases noise and introduces ambiguity into the detection, making it difficult to detect a small and fast-moving bike. In contrast, our method leverages the complementary advantages of different sensors, enabling reliable detection even under such adverse conditions.

\noindent
\textbf{Snow.} Under snowy conditions, RTNH fails to detect stationary objects due to the inherent characteristics of radar, which is more sensitive to moving targets. VoxelNeXt detects all objects, but LiDAR noise induced by snow leads to uncertain predictions, producing multiple ambiguous bounding boxes around the vehicle’s LiDAR cluster center. LoGoNet also fails to detect some objects because the RGB images are partially occluded by snow. 
In this snow scenario, thermal sensing provides more reliable cues than RGB and LiDAR, and our model effectively exploits high-confidence thermal responses while down-weighting noisy measurements from other sensors, resulting in accurate and stable detections.

\section{Sensitivity of Extrinsic Calibration} 
\label{sec:sensitivity}
We perturb the extrinsic calibrations between modalities by adding random noise to evaluate the sensitivity of our model to calibration errors. Table~\ref{tab:noise} shows that the performance decreases under noisy calibration as expected. However, the degradation is not substantial, indicating that our model remains fairly robust even under realistic levels of calibration noise. These results suggest that the model does not rely excessively on precise extrinsic calibration and can maintain stable performance in the presence of moderate calibration perturbations.

\setlength{\aboverulesep}{-1.5pt}
\setlength{\belowrulesep}{0pt}
\setlength{\tabcolsep}{9.5pt}
\renewcommand{\arraystretch}{1.00}
\begin{table}[h]
% \vspace{-10pt}
\begin{center}
\caption{Detection performance sensitivity to calibration errors.}
\vspace{-7pt}
\label{tab:noise}
\resizebox{.99\linewidth}{!}{
\begin{tabular}{c|cccccc|cccc}
% \toprule
% \multirow{2}{*}{Noise} &  \multirow{2}{*}{Clear} & \multirow{2}{*}{Fog} & Light & Heavy & Light & Heavy  & \multirow{2}{*}{Normal} & Low  & Over  & \multirow{2}{*}{HDR} \\
% & & & Rain & Rain & Snow & Snow & & Light & Expose & 
% \\
% Noise & \multirow{1}{*}{C} & \multirow{1}{*}{F} & LR & HR & LS & HS  & \multirow{1}{*}{N} & LL  & OE  & \multirow{1}{*}{HDR}
% \\
\toprule
\multirow{3}{*}{\thead{Noise \\ (translation, rotate)}}& \multicolumn{6}{c|}{Weather Condition}  & 
\multicolumn{4}{c}{Light Condition} \\
\cline{2-11}
& \multirow{2}{*}{Clear} & \multirow{2}{*}{Fog} & Light & Heavy & Light & Heavy  & \multirow{2}{*}{Normal} & Low  & Over  & \multirow{2}{*}{HDR} \\
& & & Rain & Rain & Snow & Snow & & Light & Expose & 
\\
\hline
$0$cm, $0^\circ$ & \textbf{90.30} & \textbf{71.42} & \textbf{95.10} & \textbf{80.26} & \textbf{85.59} & \textbf{72.94} & \textbf{82.93} & \textbf{92.65} & \textbf{85.47} & \textbf{86.33} \\
% $1$cm, $1^\circ$ \\
$3$cm, $3^\circ$ & \underline{89.96} & \underline{70.19} & \underline{94.60} & \underline{79.50} & \underline{84.78} & \underline{72.22} & \underline{82.51} & \underline{92.33} & \underline{84.81} & \underline{85.90} \\
$5$cm, $5^\circ$ & 88.98 & 68.46 & 93.75 & 79.42 & 83.67 & 71.72 & 82.02 & 92.02 & 84.16 & 85.75 \\
\bottomrule
\end{tabular}
}
\end{center}

% \vspace{-11pt}
\end{table}

\section{\hh{Quantitative Results for Multiple Classes}}
\label{sec:quantitative}

Following previous works~\cite{paek2022k, bijelic2020seeing, kent2024msu}, the main paper focuses on reporting results for the Vehicle category. In this supplementary material, we additionally provide results for the Pedestrian and Bike categories. For 3D object detection, Table~\ref{tab:main_3dod_1} presents the results of stereo-based and 3D range sensor–based approaches, while Table~\ref{tab:main_3dod_2} summarizes the performance of multi-modal fusion–based methods. For 2D object detection, Table~\ref{tab:main_2dod} reports the multi-class results of the methods discussed in the main paper.

\section{Acknowledgments}

This work was supported by the InnoCORE program of the Ministry of Science and ICT(N10250156), and by the Technology Innovation Program (2410013617, 20024355, Development of autonomous driving connectivity technology based on sensor-infrastructure cooperation) funded by the Ministry of Trade, Industry \& Energy(MOTIE, Korea).

\begin{table*}[p]
% \vspace{-18pt}
\centering
\caption{Comparison of datasets and benchmarks for perception in autonomous driving. The upper rows of the table list datasets built with conventional sensors, while the lower rows present datasets that include novel sensors. The symbol $\triangle$ indicates annotations that are not officially provided in the dataset but were annotated by other paper authors. If 3D bounding boxes are available, the column ‘Num Data’ reports the number of samples that include 3D bounding boxes; otherwise, it reports the total number of data samples.
}
\vspace{-3pt}
\setlength\tabcolsep{4.5pt}
\renewcommand{\arraystretch}{1.15}
\resizebox{1.0\linewidth}{!}{
\begin{tabular}{c||c|cccc|cc|ccc|ccc}
% \hline 
\thickhline 
\multirow{2}{*}{Dataset} &  
Num 
& \multicolumn{4}{c|}{Adverse Weather} & \multicolumn{2}{c|}{3D Range Sensor} 
& \multicolumn{3}{c|}{Camera Sensor}   
& \multicolumn{3}{c}{Ground-truth}   
\\
\cline{3-14}
& Data & Clear & Rain & Fog & Snow & LiDAR & Radar & Frame & Event & Thermal & 3D Bbox. & Tr. ID & Odom\\
\thickhline 
KITTI~\cite{Geiger2012AreWR} & 15k & $\tikzcmark$ & $\tikzxmark$ & $\tikzxmark$ & $\tikzxmark$ & $\tikzcmark$ & $\tikzxmark$ & Stereo & $\tikzxmark$ & $\tikzxmark$ & $\tikzcmark$ & $\tikzcmark$ & $\tikzcmark$ \\
Waymo~\cite{sun2020scalability} & 230k & $\tikzcmark$ & $\tikzcmark$ & $\tikzxmark$ & $\tikzxmark$ & $\tikzcmark$ & $\tikzxmark$ & Multi-view & $\tikzxmark$ & $\tikzxmark$ & $\tikzcmark$ & $\tikzcmark$ & $\tikzxmark$\\
NuScenes~\cite{caesar2020nuscenes} & 40k & $\tikzcmark$ & $\tikzcmark$ & $\tikzxmark$ & $\tikzxmark$ & $\tikzcmark$ & 3D & Multi-view & $\tikzxmark$ & $\tikzcmark$ & $\tikzcmark$ & $\tikzcmark$ & $\tikzcmark$\\
H3D~\cite{patil2019h3d} & 27k & $\tikzcmark$ & $\tikzxmark$ & $\tikzxmark$ & $\tikzxmark$ & $\tikzcmark$ & $\tikzxmark$ & Multi-view & $\tikzxmark$ & $\tikzxmark$ & $\tikzcmark$ & $\tikzcmark$ & $\tikzcmark$ \\
A*3D~\cite{pham20203d} & 39k & $\tikzcmark$ & $\tikzxmark$ & $\tikzcmark$ & $\tikzxmark$ & $\tikzcmark$ & $\tikzxmark$ & Stereo & $\tikzxmark$ & $\tikzxmark$ & $\tikzcmark$ &  $\tikzxmark$ & $\tikzxmark$ \\
PandaSet~\cite{xiao2021pandaset} & 8.2k & $\tikzcmark$ & $\tikzxmark$ & $\tikzxmark$ & $\tikzxmark$ & $\tikzcmark$ & $\tikzxmark$ & Multi-view & $\tikzxmark$ & $\tikzxmark$ & $\tikzcmark$ & $\tikzxmark$ & $\tikzcmark$ \\
Once~\cite{mao2021one} & 1M & $\tikzcmark$ & $\tikzcmark$ & $\tikzxmark$ & $\tikzxmark$ & $\tikzcmark$ & $\tikzxmark$ & Multi-view & $\tikzxmark$ & $\tikzxmark$ & $\tikzcmark$ & $\tikzxmark$ & $\tikzxmark$\\
Argoverse 2~\cite{wilson2023argoverse} & 150k & $\tikzcmark$ & $\tikzcmark$ & $\tikzxmark$ & $\tikzcmark$  & $\tikzcmark$ & $\tikzxmark$ & Multi-view & $\tikzxmark$ & $\tikzxmark$ & $\tikzcmark$ & $\tikzcmark$ & $\tikzcmark$ \\
CADC~\cite{pitropov2021canadian} & 8k & $\tikzcmark$ & $\tikzcmark$ & $\tikzxmark$ & $\tikzcmark$ & $\tikzcmark$ & $\tikzxmark$ & Multi-View & $\tikzxmark$ & $\tikzxmark$ & $\tikzcmark$  & $\tikzcmark$ & $\tikzcmark$ \\
Ihtaca365~\cite{diaz2022ithaca365} & 14.8k & $\tikzcmark$ & $\tikzcmark$ & $\tikzxmark$ & $\tikzcmark$ & $\tikzcmark$ & $\tikzxmark$ & Multi-view & $\tikzxmark$ & $\tikzxmark$ & $\tikzxmark$ & $\tikzxmark$ & $\tikzcmark$\\

\hline
K-Radar~\cite{paek2022k} & 35k & $\tikzcmark$ & $\tikzcmark$ & $\tikzcmark$ & $\tikzcmark$ & $\tikzcmark$ & 4D & Multi-view & $\tikzxmark$ & $\tikzxmark$ & $\tikzcmark$ & $\tikzcmark$ & $\tikzcmark$ \\
TJ4DRadSet~\cite{zheng2022tj4dradset} & 7.8k & $\tikzcmark$ & $\tikzxmark$ & $\tikzxmark$ & $\tikzxmark$ & $\tikzcmark$ & 4D & Mono & $\tikzxmark$ & $\tikzxmark$ & $\tikzcmark$  & $\tikzcmark$ & $\tikzcmark$ \\
VoD~\cite{palffy2022multi} & 8.7k & $\tikzcmark$ & $\tikzxmark$ & $\tikzxmark$ & $\tikzxmark$ & $\tikzcmark$ & 4D & Stereo & $\tikzxmark$ & $\tikzxmark$ & $\tikzcmark$  & $\tikzcmark$ & $\tikzcmark$ \\ 
RADIal~\cite{rebut2022raw} & 25k &  $\tikzcmark$ & $\tikzxmark$ & $\tikzxmark$ & $\tikzxmark$ & $\tikzcmark$ & 3D & Mono & $\tikzxmark$ & $\tikzxmark$ & $\tikzxmark$ & $\tikzxmark$ & $\tikzcmark$ \\ 
Astyx~\cite{meyer2019automotive} & 0.5k &  $\tikzcmark$ & $\tikzxmark$ & $\tikzxmark$ & $\tikzxmark$ & $\tikzcmark$ & 4D & Mono & $\tikzxmark$ & $\tikzxmark$ & $\tikzcmark$  & $\tikzxmark$ & $\tikzxmark$ \\ 
ZOD ~\cite{alibeigi2023zenseact} & 100k & $\tikzcmark$ & $\tikzcmark$ & $\tikzxmark$ & $\tikzcmark$ & $\tikzcmark$ & 4D & Mono & $\tikzxmark$ & $\tikzxmark$ & $\tikzcmark$ & $\tikzxmark$ & $\tikzcmark$ \\

SeeingThroughFog~\cite{bijelic2020seeing} & 13.5k & $\tikzcmark$ & $\tikzcmark$ & $\tikzcmark$ & $\tikzcmark$ & $\tikzcmark$ & 3D & Stereo & $\tikzxmark$ & Mono & $\tikzcmark$ & $\tikzxmark$  & $\tikzxmark$ \\
Multispectral~\cite{takumi2017multispectral} & 3.0k & $\tikzcmark$ & $\tikzcmark$ & $\tikzcmark$ & $\tikzxmark$ & $\tikzxmark$ & $\tikzxmark$ & Mono & $\tikzxmark$ & Mono & $\tikzxmark$ & $\tikzxmark$ & $\tikzxmark$ \\ 
$\text{M}^{3}\text{FD}$~\cite{liu2022target} & 4.2k & $\tikzcmark$ & $\tikzcmark$ & $\tikzcmark$ & $\tikzxmark$ & $\tikzxmark$ & $\tikzxmark$ & Mono & $\tikzxmark$ & Mono & $\tikzxmark$ & $\tikzxmark$ & $\tikzxmark$ \\ 
KAIST~\cite{choi2018kaist} & 8.9k & $\tikzcmark$ & $\tikzxmark$ & $\tikzxmark$ & $\tikzxmark$ & $\tikzcmark$ & $\tikzxmark$ & Stereo & $\tikzxmark$ & Mono & $\tikzxmark$ & $\tikzxmark$ & $\tikzxmark$ \\

DSEC~\cite{gehrig2021dsec} & 5.4k & $\tikzcmark$ & $\tikzxmark$ & $\tikzxmark$ & $\tikzxmark$ & $\tikzcmark$ & $\tikzxmark$ & Stereo & Stereo & $\tikzxmark$ & $\triangle$ & $\triangle$ & $\tikzcmark$ \\
M3ED~\cite{Chaney_2023_CVPR} & 122k & $\tikzcmark$ & $\tikzxmark$ & $\tikzxmark$ & $\tikzxmark$ & $\tikzcmark$ & \tikzxmark & Stereo & Stereo & \tikzxmark & \tikzxmark & \tikzxmark  & $\tikzcmark$ \\
\textbf{DSERT-RoLL (Ours)} & 22k & $\tikzcmark$ & $\tikzcmark$ & $\tikzcmark$ & $\tikzcmark$ & $\tikzcmark$ & 4D & Stereo & Stereo & Stereo & $\tikzcmark$ & $\tikzcmark$  & $\tikzcmark$ \\
\thickhline
% \bottomrule
\end{tabular}
}
\label{tab:dataset_compare2}
\vspace{-8pt}
\end{table*}

\setlength{\aboverulesep}{-1.5pt}
\setlength{\belowrulesep}{0pt}
\setlength{\tabcolsep}{7.2pt}
\renewcommand{\arraystretch}{1.2}
\begin{table*}[p]
\begin{center}
\caption{3D object detection performance comparison on the DSERT-RoLL dataset for vehicle (IoU = 0.5),  bike (IoU = 0.3), and  pedestrian (IoU = 0.3) detection. R, E, T, 4R, and L represent the RGB, Event, and Thermal cameras, as well as the 4D Radar and LiDAR, respectively. In addition, VEH and PED denote the vehicle and pedestrian classes, respectively. N/A indicates that the corresponding class is not present under the given adverse weather condition.}
\vspace{-7pt}
\label{tab:main_3dod_1}
\resizebox{.99\linewidth}{!}{
\begin{tabular}{c|c|c|cccccc|cccc}
\toprule
\multirow{3}{*}{Modality} & \multirow{3}{*}{Methods} & \multirow{3}{*}{Class} & \multicolumn{6}{c|}{Weather Condition}  & 
\multicolumn{4}{c}{Light Condition} \\
\cline{4-13}
&  &  & \multirow{2}{*}{Clear} & \multirow{2}{*}{Fog} & Light & Heavy & Light & Heavy  & \multirow{2}{*}{Normal} & Low  & Over  & \multirow{2}{*}{HDR} \\
& & & & & Rain & Rain & Snow & Snow & & Light & Expose & 
\\
\hline
\rowcolor{LG}
\multicolumn{13}{c}{\textbf{Stereo-based}} \\
\hline
\multirow{6}{*}{R} & \multirow{3}{*}{DSGN~\cite{chen2020dsgn}}
& VEH & 31.08 & 43.66 & 42.48 & 20.51 & 25.94 & 0.01 & 29.99 & 25.68 & 22.55 & 40.69\\
&& BIKE&  0.18 & N/A & 0.00 & N/A & N/A & N/A & 0.00 & 0.00 & N/A & 0.84\\
&& PED& 0.00 & N/A & 1.88 & N/A & 0.00 & N/A & 1.84 & 0.00 & 0.00 & 0.00\\
% \hline
\cline{2-13}
 & \multirow{3}{*}{LIGA~\cite{guo2021liga}}
 & VEH & 35.52 & 41.67 & 37.52 & 20.57 & 26.02 & 0.00 & 31.31 & 30.06 & 22.44 & 42.80\\
&& BIKE & 0.15 & N/A & 0.76 & N/A & N/A & N/A & 0.00 & 0.40 & N/A & 0.47\\
&& PED & 0.04 & N/A & 0.42 & N/A & 0.00 & N/A & 0.37 & 0.01 & 0.06 & 0.01 \\
\hline
\multirow{6}{*}{E} & \multirow{3}{*}{DSGN~\cite{chen2020dsgn}}
& VEH &  24.23 & 22.06 & 26.93 & 31.38 & 23.12 & 0.01 & 21.41 & 21.42 & 15.58 & 36.44\\
&& BIKE& 0.06 & N/A & 0.23 & N/A & N/A & N/A & 0.00 & 0.24 & N/A & 0.10\\
&& PED& 0.25 & N/A & 0.36 & N/A & 0.00 & N/A & 0.22 & 0.06 & 0.70 & 0.00\\
% \hline
\cline{2-13}
& \multirow{3}{*}{LIGA~\cite{guo2021liga}}
& VEH &  27.11 & 22.53 & 23.43 & 22.84 & 24.61 & 0.00 & 23.10 & 23.20 & 15.30 & 34.92\\
&& BIKE&0.35 & N/A & 0.11 & N/A & N/A & N/A & 0.87 & 0.07 & N/A & 0.43\\
&& PED&  0.17 & N/A & 0.74 & N/A & 0.00 &N/A & 1.07 & 0.08 & 1.01 & 0.03\\
\hline
\multirow{6}{*}{T} & \multirow{3}{*}{DSGN~\cite{chen2020dsgn}}
& VEH & 28.49 & 25.98 & 37.50 & 28.74 & 36.52 & 0.02 & 16.89 & 36.07 & 25.83 & 36.03 \\
&& BIKE&0.08 & N/A & 0.00 & N/A & N/A & N/A & 0.00 & 0.00 & N/A & 0.21\\
&& PED& 1.62 & N/A & 0.36 & N/A & 0.15 & N/A & 0.22 & 2.50 & 0.97 & 0.07\\
% \hline
\cline{2-13}
& \multirow{3}{*}{LIGA~\cite{guo2021liga}}
& VEH & 28.96 & 31.87 & 36.87 & 25.72 & 39.83 & 0.00 & 17.02 & 34.62 & 23.28 & 40.50\\
&& BIKE&1.02 & N/A & 0.00 & N/A & N/A & N/A & 0.48 & 0.02 & N/A & 2.77\\
&& PED& 0.77 & N/A & 0.84 & N/A & 1.98 & N/A & 0.55 & 1.98 & 0.50 & 0.23\\
\hline
\rowcolor{LG}
\multicolumn{13}{c}{\textbf{3D Range Sensor-based}} \\
\hline
\multirow{9}{*}{L} & \multirow{3}{*}{VoxelNeXt~\cite{chen2023voxelnext}} & VEH & 
 86.06 & 59.51 & 90.19 & 71.82 & 82.86 & 54.75 & 78.93 & 88.76 & 71.06 & 80.93\\
&& BIKE & 
 65.24 & N/A & 47.12 & N/A & N/A & N/A & 88.50 & 12.73 & N/A & 85.14
\\
&& PED & 
 26.51 & N/A & 61.14 & N/A & 72.54 & N/A & 40.15 & 33.56 & 21.39 & 15.90
\\
\cline{2-13}
& \multirow{3}{*}{HEDNet~\cite{zhang2024hednet}} & VEH & 79.27 & 48.41 & 84.74 & 68.36 & 70.29 & 55.98 & 71.64 & 83.34 & 63.97 & 73.33	\\
&& BIKE&56.76 & N/A & 4.28 & N/A & N/A & N/A & 79.59 & 17.12 & N/A & 53.86 \\
&& PED& 10.91 & N/A & 35.26 & N/A & 93.33 & N/A & 16.39 & 18.10 & 28.69 & 9.24 \\
\cline{2-13}
& \multirow{3}{*}{SAFDNet~\cite{Zhang_2024_CVPR}} & VEH & 79.30 & 43.83 & 82.82 & 57.33 & 65.07 & 49.30 & 66.28 & 81.62 & 58.38 & 76.19\\
&& BIKE&58.68 & N/A & 0.81 & N/A & N/A & N/A & 78.77 & 8.24 & N/A & 65.23\\
&& PED& 9.78 & N/A & 33.30 & N/A & 72.63 & N/A & 14.90 & 12.07 & 25.78 & 6.75 \\
\hline
\multirow{9}{*}{4R} & \multirow{3}{*}{RTNH~\cite{paek2022k}} & VEH & 23.49 & 37.30 & 43.40 & 27.86 & 36.96 & 21.70 & 28.70 & 26.28 & 24.69 & 27.00			\\
&& BIKE& 		
0.71 & N/A & 0.00 & N/A & N/A & N/A & 6.99 & 0.00 & N/A & 0.00
\\
&& PED& 0.13 & N/A & 1.36 & N/A & 1.30 & N/A & 1.39 & 0.30 & 0.47 & 0.00 \\
\cline{2-13}
& \multirow{3}{*}{VoxelNeXt~\cite{chen2023voxelnext}} & VEH & 25.03 & 44.03 & 48.78 & 27.91 & 37.42 & 32.79 & 31.82 & 24.02 & 32.50 & 35.03\\
&& BIKE& 0.24 & N/A & 0.25 & N/A & N/A & N/A & 0.03 & 1.35 & N/A & 0.00\\
&& PED& 0.16 & N/A & 4.85 & N/A & 7.93 & N/A & 0.25 & 3.71 & 0.32 & 0.00\\
\cline{2-13}
& \multirow{3}{*}{HEDNet~\cite{zhang2024hednet}} & VEH & 24.10 & 43.51 & 41.16 & 28.57 & 31.28 & 25.67 & 28.92 & 22.28 & 37.01 & 30.82\\
&& BIKE& 0.29 & N/A & 0.04 & N/A & N/A & N/A & 0.40 & 0.02 & N/A & 0.71\\
&& PED& 0.02 & N/A & 0.62 & N/A & 0.23 & N/A & 0.17 & 0.09 & 0.13 & 0.00 \\
\bottomrule
\end{tabular}
}
\end{center}
\vspace{-12pt}
\end{table*}

\setlength{\aboverulesep}{-1.5pt}
\setlength{\belowrulesep}{0pt}
\setlength{\tabcolsep}{6.2pt}
\renewcommand{\arraystretch}{1.4}
\begin{table*}[p]
\begin{center}
\caption{3D object detection performance comparison on the DSERT-RoLL dataset for vehicle (IoU = 0.5),  bike (IoU = 0.3), and  pedestrian (IoU = 0.3) detection. R, E, T, 4R, and L represent the RGB, Event, and Thermal cameras, as well as the 4D Radar and LiDAR, respectively. In addition, VEH and PED denote the vehicle and pedestrian classes, respectively. N/A indicates that the corresponding class is not present under the given adverse weather condition.}
\vspace{-7pt}
\label{tab:main_3dod_2}
\resizebox{.99\linewidth}{!}{
\begin{tabular}{c|c|c|cccccc|cccc}
\toprule
\multirow{3}{*}{Modality} & \multirow{3}{*}{Methods} & \multirow{3}{*}{Class}& \multicolumn{6}{c|}{Weather Condition}  & 
\multicolumn{4}{c}{Light Condition} \\
\cline{4-13}
&  &  & \multirow{2}{*}{Clear} & \multirow{2}{*}{Fog} & Light & Heavy & Light & Heavy  & \multirow{2}{*}{Normal} & Low  & Over  & \multirow{2}{*}{HDR} \\
& & & & & Rain & Rain & Snow & Snow & & Light & Expose & 
\\
\hline
\rowcolor{LG}
\multicolumn{13}{c}{\textbf{Multi-modal Fusion-based}} \\
\hline
\multirow{9}{*}{R+L} & \multirow{3}{*}{LoGoNet~\cite{li2023logonet}} & VEH & 
87.18 & 64.96 & 91.41 & 79.12 & 79.74 & 66.20 & 79.01 & 90.56 & 80.49 & 82.78
\\
&& BIKE&
58.52 & N/A & 2.86 & N/A & N/A & N/A & 64.87 & 10.88 & N/A & 79.29
\\
&& PED&
24.77 & N/A & 48.70 & N/A & 81.38 & N/A & 36.70 & 30.82 & 44.65 & 11.52
\\
\cline{2-13}
& \multirow{3}{*}{BEVFusion~\cite{liu2022bevfusion}} & VEH &
85.20 & 62.40 & 90.91 & 73.30 & 75.22 & 57.61 & 76.86 & 87.55 & 78.07 & 78.90
\\
&& BIKE&
52.80 & N/A & 0.00 & N/A & N/A & N/A & 72.22 & 6.10 & N/A & 67.95
\\
&& PED& 
8.99 & N/A & 32.72 & N/A & 74.84 & N/A & 17.90 & 12.75 & 32.00 & 14.72\\
\cline{2-13}
& \multirow{3}{*}{DeepFusion~\cite{li2022deepfusion}} & VEH & 87.19 & 63.94 & 91.91 & 75.61 & 81.77 & 57.26 & 79.19 & 90.81 & 78.66 & 80.10\\
&& BIKE& 53.36 & N/A & 11.55 & N/A & N/A & N/A & 77.15 & 2.55 & N/A & 69.04\\
&& PED&16.33 & N/A & 48.06 & N/A & 71.34 & N/A & 33.14 & 19.94 & 40.21 & 9.19 \\
\hline
\multirow{3}{*}{R+4R} & \multirow{3}{*}{HGSFusion~\cite{gu2025hgsfusion}} & VEH & 25.74 & 46.49 & 49.62 & 28.49 & 37.87 & 34.02 & 32.66 & 24.31 & 34.47 & 35.96\\
&& BIKE& 0.20 & N/A & 0.39 & N/A & N/A & N/A & 0.00 & 1.50 & N/A & 0.00 
\\
&& PED& 0.19 & N/A & 5.28 & N/A & 8.44 & N/A & 0.31 & 4.10 & 0.35 & 0.00 \\
\hline

\multirow{6}{*}{4R+L} & \multirow{3}{*}{InterFusion~\cite{wang2022interfusion}} & VEH & 
84.52 & 66.94 & 94.31 & 76.56 & 74.13 & 64.82 & 79.31 & 87.49 & 75.55 & 79.95\\
&& BIKE& 32.03 & N/A & 5.70 & N/A & N/A & N/A & 79.43 & 10.01 & N/A & 16.93\\
&& PED& 7.45 & N/A & 19.74 & N/A & 69.53 & N/A & 24.20 & 6.80 & 18.24 & 6.16\\
\cline{2-13}
 & \multirow{3}{*}{RL3DOD~\cite{chae2024towards}} & VEH & 
85.05 & 63.15 & 88.39 & 76.17 & 81.41 & 65.87 & 77.77 & 87.26 & 78.32 & 81.50
\\
&& BIKE&
59.85 & N/A & 13.86 & N/A & N/A & N/A & 75.51 & 13.94 & N/A & 69.73
\\
&& PED& 19.25 & N/A & 55.23 & N/A & 51.49 & N/A & 42.26 & 22.00 & 35.63 & 13.22 \\
\hline
\multirow{3}{*}{R+T+4R+L} & \multirow{3}{*}{SAMFusion~\cite{palladin2024samfusion}} & VEH & 
87.03 & 65.13 & 91.69 & 78.02 & 79.81 & 70.59 & 80.54 & 89.93 & 80.16 & 82.50
\\
&& BIKE& 
40.05 & N/A & 2.70 & N/A & N/A & N/A & 72.95 & 2.78 & N/A & 41.32
\\
&& PED& 21.09 & N/A & 43.60 & N/A & 64.34 & N/A & 38.37 & 17.49 & 41.13 & 11.51
\\
\hline
\multirow{3}{*}{R+E+T+4R+L} & 
\multirow{3}{*}{\textbf{Ours}} & VEH & 90.30 & 71.42 & 95.10 & 80.26 & 85.59 & 72.94 & 82.93 & 92.65 & 85.47 & 86.33\\
&& BIKE& 66.33 & N/A & 19.26 & N/A & N/A & N/A & 83.73 & 17.25 & N/A & 85.41\\
&& PED& 28.67 & N/A & 55.64 & N/A & 55.09 & N/A & 50.28 & 25.40 & 42.39 & 45.46\\
% \hline
\bottomrule
\end{tabular}
}
\end{center}
\vspace{-12pt}
\end{table*}

\setlength{\aboverulesep}{-1.5pt}
\setlength{\belowrulesep}{0pt}
\setlength{\tabcolsep}{6.8pt}
\renewcommand{\arraystretch}{1.4}
\begin{table*}[p]
\begin{center}
\caption{2D object detection performance comparison on the DSERT-RoLL dataset for vehicle (IoU = 0.5),  bike (IoU = 0.3), and  pedestrian (IoU = 0.3) detection. R, E, and, T represent the RGB, Event, and Thermal cameras respectively. In addition, VEH and PED denote the vehicle and pedestrian classes, respectively. N/A indicates that the corresponding class is not present under the given adverse weather condition.}
\vspace{-7pt}
\label{tab:main_2dod}
\resizebox{.99\linewidth}{!}{
\begin{tabular}{c|c|c|cccccc|cccc}
\toprule
\multirow{3}{*}{Modality} & \multirow{3}{*}{Methods} & \multirow{3}{*}{Class}& \multicolumn{6}{c|}{Weather Condition}  & 
\multicolumn{4}{c}{Light Condition} \\
\cline{4-13}
&  &  & \multirow{2}{*}{Clear} & \multirow{2}{*}{Fog} & Light & Heavy & Light & Heavy  & \multirow{2}{*}{Normal} & Low  & Over  & \multirow{2}{*}{HDR} \\
& & & & & Rain & Rain & Snow & Snow & & Light & Expose & 
\\
\hline
\rowcolor{LG}
\multicolumn{13}{c}{\textbf{Multi-modal Fusion-based}} \\
\hline
\multirow{6}{*}{R} & \multirow{3}{*}{YOLOv10~\cite{wang2024yolov10}} & VEH & 
76.47 & 72.99 & 84.95 & 76.68 & 58.76 & 2.84 & 71.98 & 67.69 & 76.25 & 76.15 
\\ 
&& BIKE & 43.13 & N/A & 24.74 & N/A & N/A & N/A & 73.50 & 10.32 & N/A & 32.76
\\
&& PED & 12.36 & N/A & 19.97 & N/A & 0.691 & N/A & 15.59 & 6.59 &  23.22 & 1.98

\\
\cline{2-13}
& \multirow{3}{*}{DEIM~\cite{huang2025deim}} & VEH &
81.85 & 82.99 & 91.48 & 73.60 & 65.07 & 13.37 & 77.76 & 72.74 & 85.14 & 79.50 \\
&& BIKE & 68.47 & N/A & 4.86 & N/A & N/A & N/A & 87.84 & 2.25 & N/A & 74.23 \\
&& PED & 22.47 & N/A & 38.59 & N/A & 13.25 & N/A & 39.90 & 19.91 & 32.79 & 25.77 \\
\hline
\multirow{6}{*}{E} & \multirow{3}{*}{RT-DETR~\cite{zhao2024detrs}} & VEH & 
73.77 & 83.17 & 83.57 & 58.93 & 47.28 & 0.023 & 69.89 & 58.28 & 78.87 & 77.83 \\
&& BIKE & 49.99 & N/A & 0.983 & N/A & N/A & N/A & 72.26 & 10.07 & N/A & 39.25 
\\
&& PED & 4.88 & N/A & 8.79 & N/A & 0.204 & N/A &  2.35 & 5.28 & 16.15 & 0.346 
\\
\cline{2-13}
& \multirow{3}{*}{DEIM~\cite{huang2025deim}} & VEH &
65.56 & 85.67 & 80.77 & 64.36 & 50.00 & 0.075 & 69.31 & 53.94 & 57.20 & 69.38 \\
&& BIKE & 63.30 & N/A &  1.30 & N/A &  N/A & N/A & 82.91 & 25.08 & N/A & 53.44 \\
&& PED & 10.52 & N/A &  2.90 & N/A & 0.05 & N/A & 4.44 & 5.19 & 22.27 & 0.837 \\
\hline
\multirow{6}{*}{T} & \multirow{3}{*}{YOLOv10~\cite{wang2024yolov10}} & VEH &
78.31 & 83.84 & 92.16 & 76.75 & 75.30 & 0.619 & 69.48 & 74.15 & 73.93 & 81.03 \\
&& BIKE & 54.29 & N/A & 78.37 & N/A & N/A & N/A & 64.29 & 59.54 & N/A & 67.50 \\
&& PED & 43.93 & N/A & 30.35 & N/A & 43.62 & N/A & 27.07 & 49.90 & 31.81 & 51.66  \\
\cline{2-13}
 & \multirow{3}{*}{DEIM~\cite{huang2025deim}} & VEH 
& 81.84 & 85.56 & 83.21 & 77.75 & 77.04 & 0.576 & 66.07 & 76.69 & 84.91 & 86.19\\
&& BIKE & 72.01 & N/A & 47.17 & N/A & N/A & N/A & 89.77 & 70.90 & N/A & 58.30 \\
&& PED & 44.38 & N/A & 27.77 & N/A & 29.43 & N/A & 20.66 & 49.47 & 23.78 & 70.42 \\
\hline
\multirow{3}{*}{R+E} & \multirow{3}{*}{GM-DETR~\cite{xiao2024gm}} & VEH 
& 84.24 & 87.54 & 95.07 & 80.92 & 59.44 & 15.62 & 83.32 & 73.61 & 86.04 & 81.90\\
&& BIKE & 78.68 & N/A & 1.94 & N/A & N/A & N/A & 93.07 & 26.03 & N/A & 72.02 \\
&& PED & 26.75 & N/A & 29.22 & N/A & 9.11 & N/A & 23.54 & 27.99 & 5.35 & 44.16 \\
\hline
\multirow{3}{*}{R+T} & \multirow{3}{*}{GM-DETR~\cite{xiao2024gm}} & VEH
& 84.10 & 86.64 & 92.18 & 77.99 & 79.44 & 1.48 & 71.87 & 77.41 & 86.12 & 88.70 \\
&& BIKE& 75.33 & N/A & 38.63 & N/A & N/A & N/A & 97.67 & 38.49 & N/A & 63.12 \\
&& PED & 42.37 & N/A & 39.68 & N/A & 43.92 & N/A & 38.36 & 44.38 & 36.67 & 76.19 \\
\hline
\multirow{3}{*}{E+T} & \multirow{3}{*}{GM-DETR~\cite{xiao2024gm}} & VEH &
85.44 & 92.13 & 88.35 & 79.64 & 81.19 & 11.20 & 71.00 & 78.96 & 87.74 & 93.04 \\
&& BIKE & 76.45 & N/A & 81.47 & N/A & N/A & N/A & 95.92 & 73.82 & N/A & 63.94 \\
&& PED  & 52.42 & N/A & 51.46 & N/A & 48.40 & N/A & 43.64 & 60.65 & 34.90 & 71.11 \\
\hline
\multirow{3}{*}{R+E+T} & \multirow{3}{*}{GM-DETR~\cite{xiao2024gm}} & VEH &
90.36 & 93.66 & 96.28 & 82.29 & 81.60 & 16.56 & 82.07 & 82.60 & 94.93 & 93.52\\
&& BIKE & 82.05 & N/A & 7.63 & N/A & N/A & N/A & 97.87 & 9.88 & N/A & 86.40 \\
&& PED & 76.90 & N/A & 61.98 & N/A & 53.41 & N/A & 64.07 & 71.53 & 72.60 & 84.96 \\
% \hline
\bottomrule
\end{tabular}
 }
\end{center}
\vspace{-12pt}
\end{table*}

\newpage

{
    \small
    \bibliographystyle{ieeenat_fullname}
    \bibliography{main}
}

\end{document}